\documentclass[journal]{IEEEtran}
\pdfoutput=1
\usepackage[table]{xcolor}
\usepackage{amsmath,amsfonts}
\usepackage{algorithmic}
\usepackage{algorithm}
\usepackage{array}
\usepackage[caption=false,font=normalsize,labelfont=sf,textfont=sf]{subfig}
\usepackage{textcomp}
\usepackage{stfloats}
\usepackage{url}
\usepackage{verbatim}
\usepackage{graphicx}
\usepackage[style=ieee, backend=biber]{biblatex}
\usepackage{booktabs}
\usepackage{multirow}
\usepackage{float}
\usepackage{tabularx}
\usepackage{hyperref}
\usepackage{caption} 
\usepackage{arydshln}

\usepackage{todonotes}

\usepackage[T1]{fontenc}
 
\usepackage[commandnameprefix=always]{changes}

\newcolumntype{C}[1]{>{\centering\arraybackslash}p{#1}}
\graphicspath{{Figures/}}
\hypersetup{hidelinks,
    	colorlinks=false,
    	linkcolor=black,
        anchorcolor=black,
        citecolor=black,
    	pdfstartview=Fit,
    	breaklinks=true}
\usepackage{titlesec} 
\titleformat{\section}[hang]{\normalsize\bfseries}{\arabic{section}}{1em}{}[]
\titleformat{\subsection}[block]{\normalsize\itshape\bfseries}{\arabic{section}.\arabic{subsection}}{1em}{}[]
\titleformat{\subsubsection}[block]{\normalsize\itshape\bfseries}{\arabic{section}.\arabic{subsection}.\arabic{subsubsection}}{1em}{}[]

\begin{document}

\title{MC-NeRF: Multi-Camera Neural Radiance Fields for Multi-Camera Image Acquisition Systems}

\author{Yu Gao$^{1,2}$, Lutong Su$^{1,2}$, Hao Liang$^{1,2}$, Yufeng Yue$^{1,2}$, Yi Yang$^{1,2 *}$, Mengyin Fu$^{1,2}$
\thanks{*This work was partly supported by National Natural Science Foundation of China (Grant No.NSFC 62233002) and National Key R\&D Program of China (2022YFC2603600)}
\thanks{$^{1}$School of Automation, Beijing Institute of Technology, Beijing, China}
\thanks{$^{2}$National Key Lab of Autonomous Intelligent Unmanned Systems, Beijing Institute of Technology, Beijing, China}
\thanks{*Corresponding author: Y. Yang Email: yang\_yi@bit.edu.cn}}
\markboth{IEEE TRANSACTIONS ON VISUALIZATION AND COMPUTER GRAPHICS, VOL. XX, NO. X, September 2023}%
{Shell \MakeLowercase{\textit{et al.}}: A Sample Article Using IEEEtran.cls for IEEE Journals}

\maketitle

\begin{abstract}
Neural Radiance Fields (NeRF) use multi-view images for 3D scene representation, demonstrating remarkable performance. As one of the primary sources of multi-view images, multi-camera systems encounter challenges such as varying intrinsic parameters and frequent pose changes. Most previous NeRF-based methods assume a unique camera and rarely consider multi-camera scenarios. Besides, some NeRF methods that can optimize intrinsic and extrinsic parameters still remain susceptible to suboptimal solutions when these parameters are poor initialized. In this paper, we propose MC-NeRF, a method for joint optimization of both intrinsic and extrinsic parameters alongside NeRF, allowing individual camera parameters for each image. First, we analyze the coupling issue that arises from the joint optimization between intrinsics and extrinsics, and propose a decoupling constraint utilizing auxiliary images. To further address the degenerate cases in the decoupling process, we introduce an efficient auxiliary image acquisition scheme to mitigate these effects. Furthermore, recognizing that most existing datasets are designed for a unique camera, we provided a new dataset that includes both simulated data and real-world data. Experiments demonstrate the effectiveness of our method in scenarios where each image corresponds to different camera parameters. Specifically, our approach outperforms the baselines favorably in terms of intrinsics estimation, extrinsics estimation, scale estimation, and rendering quality. The Code and supplementary materials are available at \href{https://in2-viaun.github.io/MC-NeRF/}{https://in2-viaun.github.io/MC-NeRF}

\end{abstract}


\section{Introduction}
Visual 3D reconstruction is pivotal in security surveillance, facial reconstruction, entertainment and robotics. These applications often involve scenes with multi-camera image
acquisition systems, providing a wealth of visual information through multi-view perspectives, such as multi-view surveillance systems and facial image capture systems.

\begin{figure}[htbp]
\centering
	\includegraphics[width=\linewidth]{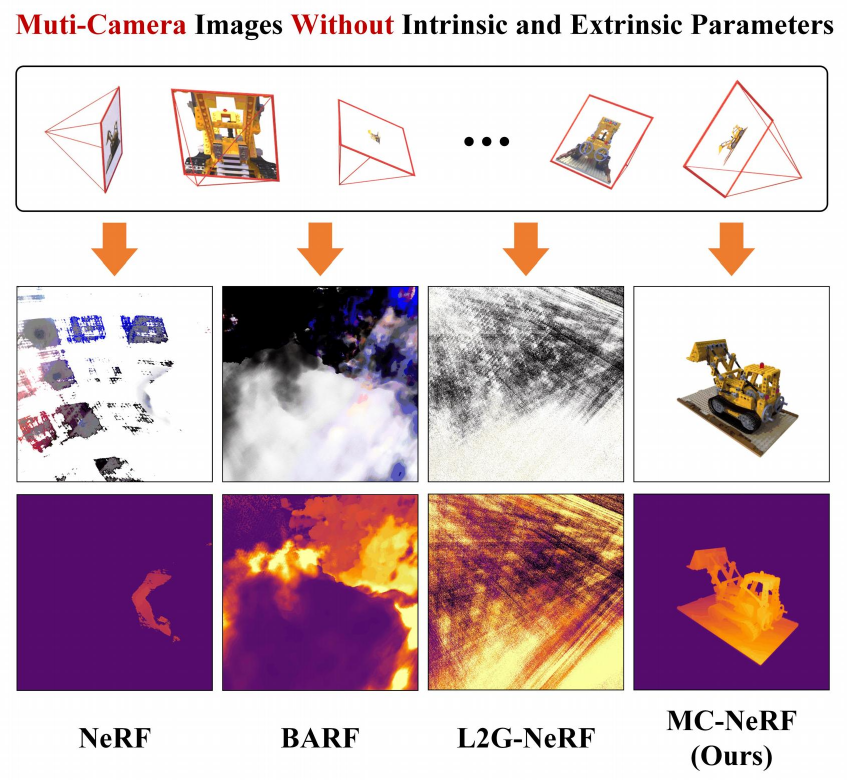}
	\caption{We introduce MC-NeRF, which can jointly optimize camera parameters with NeRF. Different from other joint optimization methods, MC-NeRF breaks the assumption of unique camera and does not require providing initial camera parameters.}
	\label{Fig.1}
\end{figure}

Recently, Neural Radiance Field (NeRF) \cite{ref1} has shown remarkable capabilities in high-quality 3D scene representation. An important precondition for NeRF series methods is the acquisition of images from different views. These images often need to be captured using a unique camera, ensuring uniform camera intrinsic parameters across all images. Specifically, widely utilized NeRF datasets, like Synthesis \cite{ref1}, LLFF \cite{ref2}, NSVF \cite{ref3},  and Mip-NeRF 360 \cite{ref4}, employ unique camera intrinsic parameters for each scene, ensuring the ray distribution model is fixed in rendering process.

However, in multi-camera image acquisition systems, ensuring that all images are captured by a unique camera is not feasible. When using images captured from these systems for reconstruction based on NeRF methods, the following challenges are encountered: Firstly, each image captured by the system corresponds to different intrinsic and extrinsic parameters, which breaks the assumption of camera uniqueness in the NeRF series methods. Secondly, accurately obtaining the parameters of each camera is time-consuming and involves a significant amount of work, and this calibration process must be repeated whenever any camera in the system changes.

A straightforward solution is using Structure-from-Motion (SfM) methods such as COLMAP \cite{ref5} to estimate camera parameters, but these techniques can be brittle and failed, especially with sparse or wide-baseline views \cite{ref55}. Moreover, the camera parameters obtained by COLMAP represent equivalent camera parameters and cannot restore the real-world scale as parameters obtained through calibration.

Another feasible solution involves NeRF methods that jointly optimize camera parameters, such as SC-NeRF \cite{ref41}, NeRF$--$ \cite{ref39}, L2G-NeRF \cite{ref54}, and BARF \cite{ref38}. However, these methods still remain susceptible to suboptimal solutions when parameters are poor initialized. Besides, these methods also follow the assumption of a unique camera, which conflicts with the first challenge. 

Simply changing the assumption from a unique camera to a multi-camera setup significantly expands the solution space of the joint optimization. This results in the principal points of the intrinsics and the translation items of the extrinsics becoming indistinguishable, preventing them from being accurately estimated. We refer to this as a coupling issue.

Focusing on the challenge, we propose Multi-Camera NeRF (MC-NeRF), a novel 3D reconstruction method for multi-camera image acquisition systems. We establish a global relationship among cameras by introducing additional constraints through the reprojection process. On one hand, the reprojection loss ensures a stable and unique solution space, effectively resolving the coupling issue. On the other hand, given a unique solution for the intrinsics, optimizing the reprojection loss through bundle adjustment (BA) can serve as a method for initializing the extrinsics. Additionally, we also found that not all reprojection processes in a scene can provide effective constraints. We define the cases where they fail as degeneration cases and propose an efficient auxiliary image acquisition scheme to avoid these situations.

As for the datasets, the majority of existing NeRF datasets are generated based on a unique camera, which can not satisfy the requirement of randomized camera parameters with mixed cameras. This has motivated us to propose our own dataset. The MC dataset provides the reader with the flexibility to tailor camera parameters and the number of cameras, allowing for free combinations. Furthermore, to demonstrate the effectiveness of our method in real-world applications, we constructed a multi-camera image acquisition system comprising 88 cameras and provided a dataset from real-world scenarios.

In conclusion, the contributions of this paper are as follows:

$\bullet$ We address the coupling issue within joint optimization of intrinsic and extrinsic parameters and the degenerated cases in intrinsic parameter estimation.

$\bullet$ We propose a joint optimization strategy that can simultaneously optimize the intrinsic parameters, extrinsic parameters and neural radiation fields.

$\bullet$ We construct a multi-camera image acquisition system comprising 88 cameras and provided a new dataset of real-world scenes captured by this system.

$\bullet$ We provide a new simulated dataset for multi-camera acquisition systems including the source code for dataset generation, enabling readers to create their own datasets freely.

\begin{figure*}[htbp]
\centering
	\includegraphics[width=\linewidth]{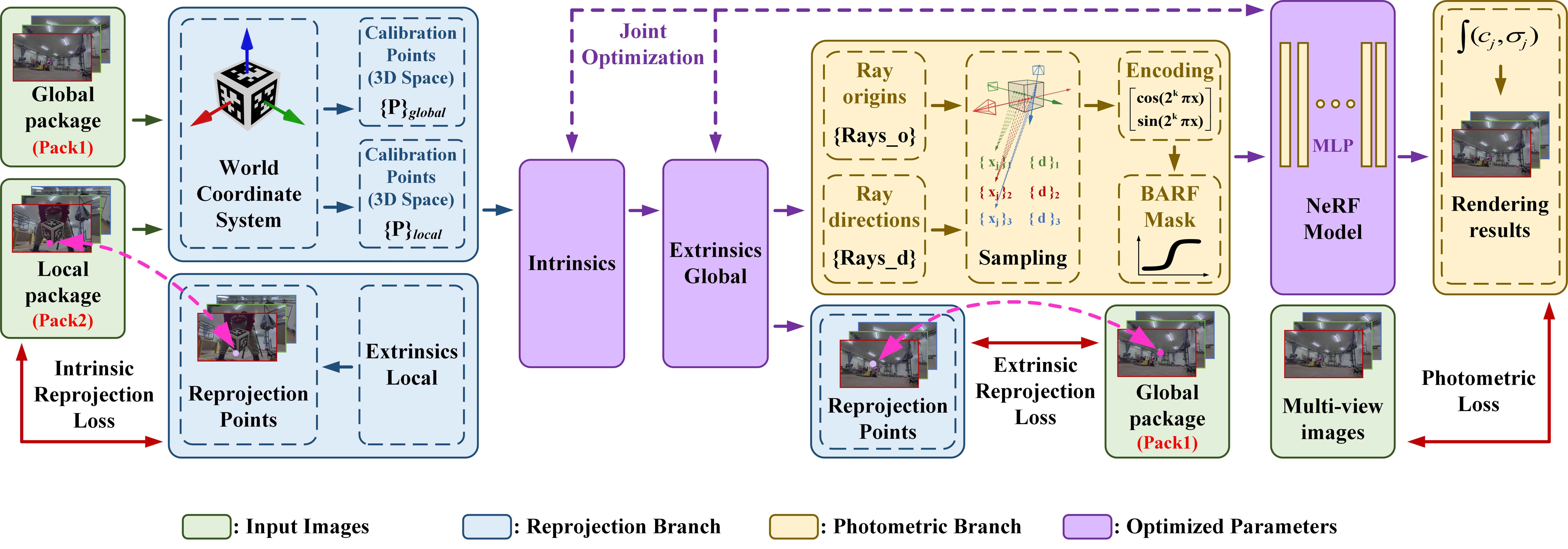}
	\caption{Overview of the proposed method. MC-NeRF utilizes three sets of images as inputs. $Pack1$ is used to unify the world coordinate systems across all cameras and provide initial extrinsics. $Pack2$ imposes intrinsic constraints to address the coupling issue that arises during joint optimization. Multi-view images are scene images, which are used for reconstructing the 3D scene. $Pack1$ and $Pack2$ provide 3D coordinates in space and feature points in the images, which are utilized to establish reprojection loss functions. Then, the intrinsics and extrinsics constrained by the reprojection loss function are used to generate sampling points. These points are subsequently fed into a Multilayer Perceptron (MLP) that employs BARF-based progressive encoding for further training. To ensure both efficiency and convergence, the training sequence must follow the process outlined in Fig.\hyperref[Fig.6]{\ref{Fig.6}}}
	\label{Fig.2}
\end{figure*}

\section{RELATED WORK}
\noindent \textbf{Camera Calibration} \quad 
The camera model demonstrates how points in camera coordinates are projected into pixel coordinates, and camera calibration aims to determine the parameters within this model. Earlier calibration methods observe calibration objects whose geometry in 3D space is known with high precision to calculate the camera parameters, and this efficient method requires only a small number of images to achieve high-performance results \cite{ref8}. Considering the precision of manufacturing and the cost of calibration objects, Tsai et al. \cite{ref9} captured images of a 2D planar calibration object from various poses to estimate camera parameters. This method requires providing accurate pose for each calibration plane. Another widely used approach adopting 2D planar calibration is the chessboard method proposed by Zhang \cite{ref10}. This technique involves capturing images including a chessboard pattern from different positions to determine the camera parameters. Compared with Tsai's method, Zhang's approach eliminates the need to provide the calibrator's position. In addition to employing known calibration points, there are also some methods relying on detecting specific features to achieve calibration, like vanishing points \cite{ref11,ref12} and coplanar circles \cite{ref13}. 

In recent years, deep learning has provided new inspirations for camera calibration. DeepCalib \cite{ref14} utilizes a CNN network trained with a vast collection of scene images matched with corresponding camera parameters, and established the relationship between image features and camera parameters. DeepCalib can directly predict calibration results based on the input images. An advantage of this method is that there is no need to provide typical calibration objects. Similar to DeepCalib, Hold-Geoffroy et al. \cite{ref15} also attempted to predict camera parameters directly from a single image, and they explored the boundaries of human tolerance towards image distortion, and further designed a new perceptual measure for the calibration errors.

Apart from methods designed explicitly to obtain calibration results, some works can indirectly acquire camera parameters. Some learning methods \cite{ref16,ref20} rely on classical multi-view geometry theories, like epipolar constraints, PnP (Perspective-n-Point) algorithms, feature points matching and triangulation, to achieve 3D reconstruction or visual odometry. Although these applications may not design for obtaining camera parameters, they still has the capability to retrieve poses, and the mentioned theories always serve as the foundation for model block design and loss functions design. Additionally, camera pose estimation methods based on NeRF, such as iNeRF \cite{ref24}, have also emerged.

\noindent \textbf{Camera Preconditions of NeRF and 3D Gaussian} \quad
Acquiring accurate camera parameters can often be challenging, and in some cases, certain parameters might even be unreachable. When camera parameters are unreliable, adopting classic 3D Gaussian \cite{ref56} and NeRF-series methods \cite{ref26,ref31,ref32} to describe the scene becomes difficult. Some researches \cite{ref34,ref35,ref36} employ input images to estimate camera poses. Li et al. \cite{ref26} used COLMAP to obtain the initial camera poses, which were subsequently used as precondition for NeRF to reconstruct the scene. Instead of using existing methods directly, some other methods \cite{ref37,ref40} can leverage prior knowledge to establish the loss function by introducing geometric constraints with photometric consistency. Specifically, Wang et al. \cite{ref39} jointly optimized camera parameters and NeRF through photometric reconstruction. Jeong et al \cite{ref41} design a mixed camera model to achieve the self-calibration, and the process is constrained by geometric and photometric consistency for NeRF training. Lin et al \cite{ref38} employed a coarse-to-fine auxiliary positional embedding \cite{ref42,ref43,ref44} to jointly recover the radiance field and camera pose. NoPe-NeRF \cite{ref45} uses monocular depth priors to constrain the scene as well as relative pose estimates.  Ray-to-ray correspondence losses also can be leveraged to impose constraints on the relative pose estimates by utilizing either keypoint matches or dense correspondences \cite{ref41,ref46}. DBARF \cite{ref47} adopts low-frequency feature maps to guide the bundle adjustment for generalizable NeRFs \cite{ref48,ref49}. GNeRF \cite{ref50} introduces a pose estimator to directly estimate camera poses from images. CAMP \cite{ref55} proposes a preconditioner to balance the influence of items in the camera parameters during the joint optimization process, thereby enhancing the quality of the reconstruction. Essentially, these methods combine the training process with camera parameters prediction.

Compared to NeRF, the joint optimization of 3D Gaussians and camera parameters is more complex. Due to the anisotropic characteristics of 3D Gaussian primitives, existing methods such as InstantSplat \cite{ref57}, CF-GS \cite{ref58}, GS-SLAM \cite{ref59}, and SplaTAM \cite{ref60} adopt a fixed camera pose and rotate the scene to obtain the inverse extrinsics, which is then transformed back to the extrinsics. Among these methods, InstantSplat is capable of optimizing both intrinsics and extrinsics simultaneously. It utilizes DUSt3R \cite{ref61} to obtain initial point clouds and camera parameters, and the process of converting intrinsics and extrinsics into learnable parameters is consistent with that of NeRF$--$. Essentially, the idea of these 3D Gaussian methods is to find an optimal combination of camera parameters that renders images with quality consistent with the ground truth images.

In summary, NeRF and 3D Gaussian methods have been extensively explored and significantly improved, yet certain limitations remain. Firstly, previous works often require inputs collected from the same camera or cameras with identical performance, which is difficult to ensure in a multi-camera image acquisition system. Secondly, different intrinsic parameters are rarely considered in prior researches, implying that existing methods aim to standardize all images to a single camera model with uniform parameters.

\section{METHOD}
\subsection{Overview}

The proposed method, illustrated in Fig.\hyperref[Fig.2]{\ref{Fig.2}}. consists of three loss functions and is divided into two branches. It utilizes three sets of images as inputs and does not require any camera initialization parameters. The first set $I$ consists of scene or object images captured by a multi-view system, the second set ${P_g}$ includes global auxiliary images, and the third set ${L_g}$ contains local auxiliary images, as indicated by the green boxes in Fig.\hyperref[Fig.2]{\ref{Fig.2}}. The method can be described as follows:

\begin{equation}
    Scene,Intrs,Extrs = MC-NeRF(I,{P_g},{L_g})
\label{eq:1}
\end{equation}

$Scene$ represents the reconstructed scene, while $Intrs$ and $Extrs$ denote the intrinsics and extrinsics of the camera corresponding to the current train image. Since the auxiliary images contain real-world size of AprilTags, the reconstructed scene from MC-NeRF is at real-world scale, which differs from the equivalent scene reconstructed using SFM parameters.

In the following sections, we first address the coupling issues of intrinsics and extrinsics in joint optimization and introduce our solution (Sec 3.2). Next, we discuss the failure cases of the solution, referred to as the degenerate case, and propose an efficient auxiliary image acquisition scheme to avoid this case (Sec 3.3). Following this, we present the training details of our method, including parameterization, loss functions, and the training schedule (Sec 3.4). Finally, we introduce the multi-camera system we used and the public dataset we released (Sec 3.5).

\subsection{Coupling Issue in Joint Optimization} 

We unfold this section by explaining the coupling issue (Sec 3.2.1). We then explore the necessary constraints for jointly optimizing camera parameters and NeRF with existing methods (Sec 3.2.2). Following this, we introduce a solution to the coupling issue when each camera has its own individual intrinsics and extrinsics (Sec 3.2.3).

\subsubsection{Coupling in camera parameters}
During the joint optimization, $MLP$, $[R|T]$, and $K$ are defined as learnable variables. $[R|T]$ is the camera extrinsics, and $K$ is the camera intrinsics. $enc$ represents the position encoding function, and the execution workflow of NeRF can be described as follows: 

\begin{equation}  
\begin{aligned} 
    RGB,\sigma &= MLP\left( {enc(\{ {S_n}\}, \{ {D_n}\} )} \right)
    \quad\\
   {\text{      with }} \left\{ {{S_n}} \right\} &= {f_{sample}}(\left[ {R|T} \right],K,pt{s_n})
\end{aligned} 
\label{eq:2}  
\end{equation}

$RGB$ and $\sigma$ denote the color and density of the sample points $\left\{ {S_n} \right\}$ in 3D space. $\left\{ {D_n} \right\}$ represents the view direction. $f_{sample}$ is the function that converts pixel coordinates into sampled points, and $pts_n$ is the pixel coordinate of pixel $n$. Specifically, $f_{sample}$ can be rewritten as follows:

\begin{equation}
\left\{ {{S_n}} \right\} = ray_o + ray_d \cdot \left\{ {{s_n}} \right\}
\label{eq:3}  
\end{equation}

$\left\{ {s_n} \right\}$ represents the sample points in standard space, which are related to the sampling distance. $ray_o$ denotes the origin and $ray_d$ represents the direction vector of the emitted ray. These items can be obtained as:

\begin{equation}  
\begin{aligned} 
    {ray_o} &= T
    \quad\\
    {ray_d} &= norm(Pts_n - T)
    \quad\\
    {Pts_n} &= {[R|T]^{ - 1}}{K^{ - 1}}{pts_n}
\end{aligned} 
\label{eq:4}  
\end{equation}

 $Pts_n$ is 3D coordinate of $pts_n$ projected into 3D space. $norm$ represents the normalization operation. Moreover, ${Pts_n}$ in Eq.\eqref{eq:4} can be modified as:

\begin{equation}
\begin{aligned}
{Pts_n} &= {[R|T]^{ - 1}}{K^{ - 1}}{pts_n}\\
\Rightarrow{Pts_n} &= {R^{ - 1}}{K^{ - 1}}{pts_n} - {R^{ - 1}}T\\
\Rightarrow{Pts_n} &= \tilde R{pts_n}{\rm{ + (}}\tilde T - {R^{ - 1}}T{\rm{)}}
\end{aligned} 
\label{eq:5}
\end{equation}

We define $[R|T]$ and $K$ as follows:

\begin{equation}  
\begin{aligned} 
    {[R|T]} &=\begin{bmatrix}
    r_{11} & r_{21} & r_{31} & t_{1}\\ 
    r_{12} & r_{22} & r_{32} & t_{2}\\ 
    r_{13} & r_{23} & r_{33} & t_{3}
    \end{bmatrix},\\
    \quad\\
    K &=\begin{bmatrix}
    f_{x} & 0 & u_{0}\\ 
    0 & f_{y} & v_{0}\\ 
    0 & 0 & 1
    \end{bmatrix}
\end{aligned} 
\label{eq:6}  
\end{equation}

where $\tilde R$ and $\tilde T$ in Eq.\eqref{eq:5} are as follows:

\begin{equation}
\begin{aligned}
    & \tilde{R}=\begin{bmatrix}
r_{11}/f_{x} & r_{12}/f_{y} & r_{13}\\ 
r_{21}/f_{x} & r_{22}/f_{y} & r_{23}\\ 
r_{31}/f_{x} & r_{32}/f_{y} & r_{33}
\end{bmatrix},\\
\quad \\
& \tilde{T}=\begin{bmatrix}
-u_{0}r_{11}/f_{x}-v_{0}r_{12}/f_{y}\\ 
-u_{0}r_{21}/f_{x}-v_{0}r_{22}/f_{y}\\ 
-u_{0}r_{31}/f_{x}-v_{0}r_{32}/f_{y}
\end{bmatrix}
\end{aligned}
\label{eq:7}
\end{equation}

we also have:   
\begin{equation}
\begin{aligned}
\tilde{T}-R^{-1}T &=\begin{bmatrix}
\tilde{t}_{1}r_{11}+\tilde{t}_{2}r_{12}+\tilde{t}_{2}r_{13}\\ 
\tilde{t}_{1}r_{21}+\tilde{t}_{2}r_{22}+\tilde{t}_{2}r_{23}\\ 
\tilde{t}_{1}r_{31}+\tilde{t}_{2}r_{32}+\tilde{t}_{2}r_{33}
\end{bmatrix} \\
\quad\\
{\text{  with   }}\tilde{t}_{1} &=-u_{0}/f_{x}-t_{1}\\
\tilde{t}_{2} &=-v_{0}/f_{y}-t_{2}\\
\tilde{t}_{3} &=-t_{3}
\end{aligned}
\label{eq:8}
\end{equation}

From the last line of Eq.\eqref{eq:5}, we can observe that the 2D-to-3D projection can be equated to operations based on $\tilde R$ and $(\tilde{T}-R^{-1}T)$. From Eq.\eqref{eq:7}, $\tilde R$ can be decomposed into a combination of scaling and rotation, which are decoupled.

However, for $(\tilde{T}-R^{-1}T)$, this item represents a translation operation, which is influenced by $R$, $K$ and $T$.
Taking the first row of Eq.\eqref{eq:8} as an example, we have:

\begin{equation}
\begin{aligned}
    &\tilde{t}_{1}r_{11}+\tilde{t}_{2}r_{12}+\tilde{t}_{2}r_{13} \\
    \Rightarrow \alpha {u_0} + \beta &{v_0} + {r_{11}}{t_1} + {r_{12}}{t_2} + {r_{13}}{t_3} \\
    \quad\\
    {\text{  with   }}&\alpha = {{ - {r_{11}}} \mathord{\left/
     {\vphantom {{ - {r_{11}}} {f_x}}} \right.
     \kern-\nulldelimiterspace} {f_x}}, \beta = {{ - {r_{12}}} \mathord{\left/{\vphantom {{ - {r_{12}}} {{f_y}}}} \right.
     \kern-\nulldelimiterspace} {{f_y}}}\\
\end{aligned} 
\label{eq:9}
\end{equation}

Since rotation and scaling are decoupled, we assume that we can obtain the correct optimization results for $R$ and $f_x$, $f_y$. This means that Eq.\eqref{eq:9} contains five unknown variables: $u_0$, $v_0$ and $t_1$, $t_2$, $t_3$. According to Eq.\eqref{eq:8}, there are three constraint equations for the five parameters mentioned above, which indicates that it is not possible to accurately solve for each parameter. While we can obtain result for $(\tilde T - {R^{ - 1}}T)$ through optimization, distinguishing between $u_0$, $v_0$ and $T$ remains challenging. 

In summary, $u_0$, $v_0$, and $T$ are coupled during the joint optimization of both intrinsics and extrinsics, making it impossible to obtain accurate values for each. As a result, we can achieve good rendering performance from the view of the training images, but rendering errors occur when generating new viewpoints.

\subsubsection{Coupling in multi-camera systems}

It is important to note that in Sec 3.2.1, we considered only a case of single image and a single camera. Next, we analyze the case of multiple images and multiple cameras. The existing methods can be categorized into the following three cases:

1)\quad With intrinsics provided, joint optimization of extrinsics and NeRF, such as BARF, L2G-NeRF.

2)\quad With global unique intrinsics, joint optimization of intrinsics, extrinsics and NeRF, such as SC-NeRF, NeRF$--$.

3)\quad Joint optimization of intrinsics, extrinsics and NeRF, our goal and method.

Based on Eq.\eqref{eq:8} and Eq.\eqref{eq:9}, $(\tilde{T}-R^{-1}T)$ have the following expression:

\begin{equation}
\begin{aligned}
\Rightarrow &\begin{bmatrix}
    {\alpha _1}{u_0} + {\beta _1}{v_0} + {r_{11}}{t_1} + {r_{12}}{t_2} + {r_{13}}{t_3}\\ 
    {\alpha _2}{u_0} + {\beta _2}{v_0} + {r_{21}}{t_1} + {r_{22}}{t_2} + {r_{23}}{t_3}\\ 
    {\alpha _3}{u_0} + {\beta _3}{v_0} + {r_{31}}{t_1} + {r_{32}}{t_2} + {r_{33}}{t_3}\\
    \end{bmatrix} \\
    \quad\\
&{\text{  with   }}\alpha_1 = {{ - {r_{11}}} \mathord{\left/
     {\vphantom {{ - {r_{11}}} {f_x}}} \right.
     \kern-\nulldelimiterspace} {f_x}}, \beta_1 = {{ - {r_{12}}} \mathord{\left/{\vphantom {{ - {r_{12}}} {{f_y}}}} \right.
     \kern-\nulldelimiterspace} {{f_y}}}\\
&{\text{ \qquad }}\alpha_2 = {{ - {r_{21}}} \mathord{\left/
     {\vphantom {{ - {r_{21}}} {f_x}}} \right.
     \kern-\nulldelimiterspace} {f_x}}, \beta_2 = {{ - {r_{22}}} \mathord{\left/{\vphantom {{ - {r_{22}}} {{f_y}}}} \right.
     \kern-\nulldelimiterspace} {{f_y}}}\\
&{\text{ \qquad }}\alpha_3 = {{ - {r_{31}}} \mathord{\left/
     {\vphantom {{ - {r_{31}}} {f_x}}} \right.
     \kern-\nulldelimiterspace} {f_x}}, \beta_3 = {{ - {r_{32}}} \mathord{\left/{\vphantom {{ - {r_{32}}} {{f_y}}}} \right.
     \kern-\nulldelimiterspace} {{f_y}}}
\end{aligned}
\label{eq:10}
\end{equation}

For the first case, when the intrinsics are provided, the number of unknown variables is reduced from five to three (only $t_1$, $t_2$, $t_3$), matching the number of equation constraints in Eq.\eqref{eq:10}, allowing for joint solving. This indicates that the extrinsics can be obtained without any additional constraints, which is consistent with the BARF and L2G works. In these methods, no additional loss functions were introduced apart from those used for NeRF training.

For the second case, when the intrinsics are globally unique, the number of unknown variables is remaining five. However, $u_0$ and $v_0$ can be shared across multiple views, which means that accurate parameters can be obtained by adding only a few constraints. Assuming that we have $m$ viewpoints, and there are a total of $3m$ constraint equations, while the number of unknown parameters  is $3m+2$. For example, in SC-NeRF, the researchers introduced the Projected Ray Distance (PRD) loss function to achieve joint optimization. NeRF$--$ provides no additional constraints, but the authors emphasized the limitation of their method, which is only effective for forward-facing scenes. We believe this effectiveness benefits from the neural network's strong approximation capabilities, but this ability cannot generalize well in the absence of sufficient constraints.

For the third case, there is almost no difference between multi-view and single-view. Each viewpoint has its own intrinsics and extrinsics. Assuming we have $m$ viewpoints, there are a total of $3m$ constraint equations, while the number of unknown parameters is $5m$. This means that at least two additional constraint equations must be added for each viewpoint to obtain accurate parameters.

\begin{figure}[htbp]
	\centering
	\includegraphics[width=\linewidth]{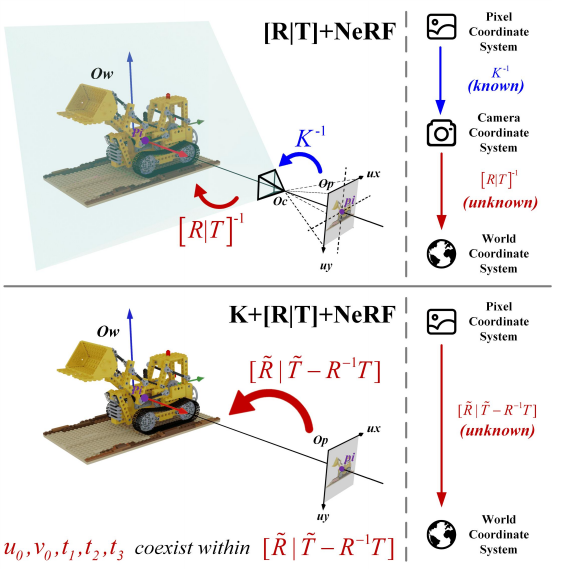}
	\caption{Coupling issue between intrinsic and extrinsic parameters. 1) The first row illustrates the joint optimization for extrinsics and NeRF. In methods such as BARF, L2G-NeRF, where the intrinsic parameters are known, mitigating the issue of camera parameters coupling. 2) The second row showcases the joint optimization of both intrinsics and extrinsic, as outlined in Eq.\eqref{eq:5}. This procedure effectively represents a transformation involving scaling, rotation, and translation. $u_0$ and $v_0$ from intrinsics, along with  $t_1$, $t_2$ and $t_3$ from extrinsics, coexist within $(\tilde T - {R^{ - 1}}T)$ and cannot be disentangled.}
	\label{Fig.3}
\end{figure}

Fig.\hyperref[Fig.3]{\ref{Fig.3}} illustrates the comparison between joint optimization of extrinsics alone and the joint optimization of both intrinsics and extrinsics, highlighting the differences. 

In summary, joint optimization in multi-camera systems is not merely a matter of turning camera parameters into learnable variables and combining them with NeRF for backpropagation training. Depending on the assumptions of the camera parameters, different numbers of constraint equations may be required to ensure the feasibility of the joint optimization results.

\subsubsection{Solution to the Coupling Issue}

Based on Sec 3.2.2, a straightforward solution to address the coupling issue is imposing additional constraints on the coupled variables. We introduce auxiliary images based constraints to separate $u_0$, $v_0$ and $t_1$, $t_2$, $t_3$ from each other. 

Given a set of 3D points $\left\{ {C_i} \right\}$ in world coordinate system, and knowing their corresponding coordinates $\left\{ {c_i} \right\}$ in pixel coordinate system. We define a homography $H$ with size $3\times 4$, which is also called camera matrix, the transformation can be defined as: 

\begin{equation}
    \left\{ {c_i} \right\}=H\left\{ {C_i} \right\} \quad \text{       with }H=K\left[ R\left| T \right. \right]
    \label{eq:11}
\end{equation}

With the condition of $i \ge 6$ and ${{\left\| H \right\|}_{F}}=1$, where $F$ is Frobenius norm, $H$ can be solved by SVD decomposition.
When $H$ is obtained, $K$ and $R$ can be solved by the first three columns of $H$. 

Eq.\eqref{eq:11} implies that if we can provide a sufficient number ($i \ge 6$) of 3D-2D correspondences, additional constraints on the intrinsics can be introduced, enabling their estimation. It is important to note that the coordinate system of 3D points provided may differ from that of the target scene requiring reconstruction. In such cases, the estimated extrinsics become meaningless, but the intrinsics can still be useful and shared across different coordinate systems.

The correspondences provides additional constraints on $u_0$ and $v_0$ in the intrinsics, ensuring that Eq.\eqref{eq:10} can be solved, resolving the coupling issue.

\begin{figure}[htbp]
	\centering
	\includegraphics[width=\linewidth]{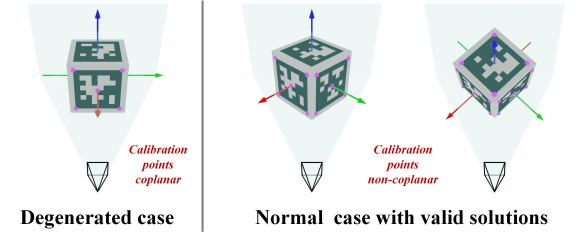}
	\caption{Degenerate case in intrinsic parameters estimation. When a single AprilTag is present in the calibration image, obtaining a valid solution is not feasible. At least two AprilTags ensures the acquisition of camera intrinsic parameters.}
	\label{Fig.4}
\end{figure}

\subsection{Degenerate Case and Auxiliary Image Scheme} 

In this section, we first explain the degenerate case and discuss how to prevent it (Sec 3.3.1). Based on this principle, we propose an efficient auxiliary image acquisition scheme (Sec 3.3.2). By using auxiliary images, we can simultaneously resolve parameter coupling and avoid degeneration during the joint optimization.

\subsubsection{Degeneration in Coupling Solution}

As discussed in Sec 3.2.3, providing more than six ($i \ge 6$) 3D-2D correspondences for each camera theoretically resolves the coupling issue. However, during practical system validation, we found that these correspondences can not always guarantee valid constraints. According to \cite{ref52}, when the 3D points in space are distributed on the same plane, Eq.\eqref{eq:11} only yields a general solution rather than a unique one. This circumstance is referred to as degeneration.

The degeneration imposes rules on the provided correspondences, requiring that these matches follow the principle of non-coplanar distribution. In classical calibration methods, such as the chessboard method, all corners lie on the same plane, which can not meet our requirements. To obtain matches that satisfy these conditions, we designed an AprilTag cube and developed an auxiliary image acquisition scheme.

\subsubsection{Auxiliary Images Capture Scheme}

We use a multi-camera system to capture images containing the AprilTag cube, providing the aforementioned 3D-2D matches. The cube has each face spray-painted with a different AprilTag label from the 36H11 family. AprilTag can provide five calibration points per plane, including four corners and a central point. Detecting more than two AprilTag patterns in each image can provide at least 10 points, which fulfills the condition of being non-coplanar. Besides, the cube is easy to machine, with low cost but high machining accuracy. As for calibration points detection, AprilTag supports an open source algorithm, which is stable and easy to deploy. 

\begin{figure}[htbp]
	\centering
	\includegraphics[width=\linewidth]{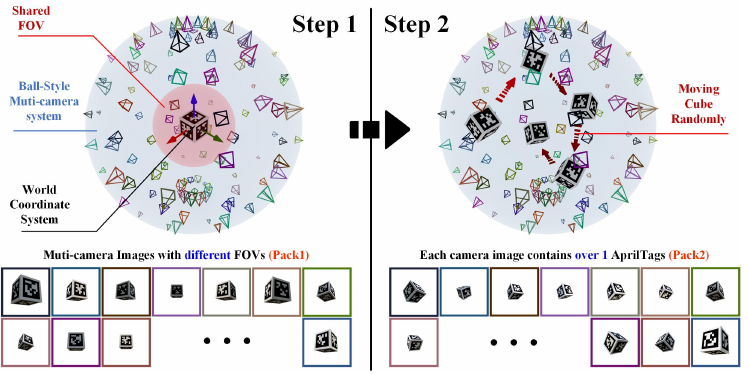}
	\caption{Details of calibration data Acquisition. Firstly, the cube is captured by all cameras within the shared field of view, with its center defined as the origin of the world coordinate system. We define the set of images captured during this step as $Pack1$. Secondly, to ensure that each camera captures at least two AprilTags, providing non-coplanar correspondences, the operator needs to randomly move the cube in front of each camera. Once a camera detects more than two AprilTags, the image can be saved. Each camera only needs to capture one such image. The set of images captured in this phase is defined as $Pack2$.}
	\label{Fig.5}
\end{figure}

Fig.\hyperref[Fig.5]{\ref{Fig.5}} illustrates the procedure for acquiring auxiliary images, and two sets of auxiliary images need to be captured. The first set, referred to as $Pack1$, is used to unify the
world coordinate systems across all cameras and provide initial extrinsics. The second set, referred to as $Pack2$, introduces additional constraints to resolve the coupling issue and avoid degenerate case. Since the necessity of the second set has already been addressed, we here focus on discussing the role and importance of the first set of images.

According to Eq.\eqref{eq:10}, after decoupling the camera parameters using the auxiliary images from $Pack2$, our challenges shift from the third case in Sec 3.2.2 to the first one. At this point, we encounter the same challenges as those presented in BARF and L2G-NeRF. Although these methods can optimize imperfect extrinsics with NeRF, they are still susceptible to suboptimal solutions when the parameters are poorly initialized, as demonstrated in Sec 4.3. To avoid these suboptimal solutions that lead to optimization failure, we utilize more robust constraints provided by $Pack1$, ensuring that our method is not compromised by initialization.

Additionally, our image acquisition scheme is highly adaptable to multi-camera systems that undergo frequent changes. If the position or orientation of any camera changes, we simply need to place the cube at the center of the area and recapture $Pack1$, while $Pack2$ can reuse previously captured results. When adding a new camera or replacing an existing one, a single image containing at least two AprilTags must be captured for each new camera, followed by the recapture of $Pack1$. We believe this manual effort is significantly less than that required for performing a complete system calibration and does not necessitate evaluating the quality of any calibration results.

\subsection{Parameters Design and Training Details} 

In this chapter, we first introduce how the camera parameters are transformed into learnable variables and outline the loss functions used in our method (Sec 3.4.1). Next, we define the world coordinate system and explain its relationship with the two sets of auxiliary images (Sec 3.4.2). Finally, given the different convergence rates of camera parameter estimation and NeRF training, we propose a training sequential diagram to ensure the efficiency of the training process (Sec 3.4.3).

\subsubsection{Parametrization and Loss Function}

\begin{figure*}[htbp]
	\centering
	\includegraphics[width=\linewidth]{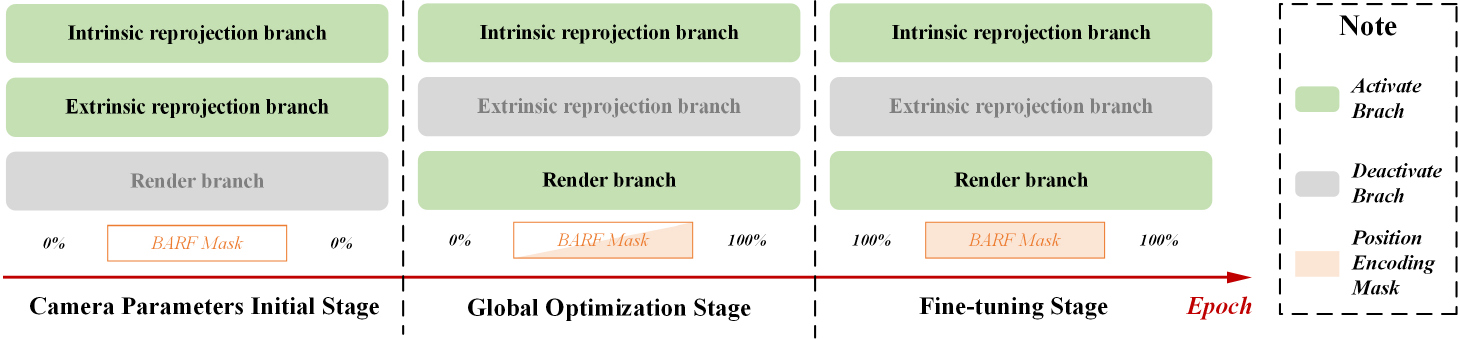}
	\caption{Training sequence diagram: Our training process consists of three stages: 1) Camera Parameter Initialization Stage: The goal in this stage is to obtain coarse intrinsic and extrinsic parameters. Rendering training is not conducted in this stage. 2) Global Optimization Stage: Due to the coupling issue during joint optimization, we retain the Intrinsic Reprojection branch as a decoupling constraint. We also employ the progressive alignment method proposed in BARF to achieve extrinsic parameters with higher accuracy than those obtained in the first stage. 3) Fine-tuning Stage: In this stage, continuing to optimize extrinsic parameters without the progressive alignment constraint can lead to divergence. We freeze the extrinsic parameters and only adjust the intrinsic parameters.}
    \label{Fig.6}
\end{figure*}

Assume that the multi-camera system contains $m$ cameras with unknown parameters, and ${{K}_{j}}$$(j=0,1,2,...,m)$ denotes the intrinsic matrix of each camera. We decompose ${{K}_{j}}$ into the initialization ${{K}_{j0}}$ and the adjustable weights $\Delta {{K}_{j0}}$, and also assume the axes of the image coordinate system are all perpendicular to each other:

\begin{equation}
    {{K}_{j}}=\left[ \begin{matrix}
    {{f}_{xj}}\cdot \Delta {{f}_{xj}} & 0 & {{u}_{0j}}\cdot \Delta {{u}_{0j}}  \\
    0 & {{f}_{yj}}\cdot \Delta {{f}_{yj}} & {{v}_{0j}}\cdot \Delta {{v}_{0j}}  \\
    0 & 0 & 1  \\
\end{matrix} \right]
\label{eq:12}
\end{equation}

We define the adjustable extrinsic parameters in the $se(3)$ space and then convert back to the $SE(3)$ space. Let $\left( {{\alpha }_{j0}},{{\alpha }_{j1}},{{\alpha }_{j2}},{{\alpha }_{j3}},{{\alpha }_{j4}},{{\alpha }_{j5}} \right)$ be 6 adjustable parameters in the $se(3)$ and the corresponding matrix in the $SE(3)$ is denoted as $\left[ {{R}_{j}}\left| {{T}_{j}} \right. \right]$ with size $3\times 4$. For every camera in the system, we rewrite Eq.\eqref{eq:11} as:

\begin{equation}
    pd\_{{c}_{ij}}={{K}_{j}}\left[ {{R}_{j}}\left| {{T}_{j}} \right. \right]{{C}_{ij}}
\label{eq:13}
\end{equation}

where ${{C}_{ij}}$ is the 3D point provided by $Pack2$, $pd\_{{c}_{ij}}$ is the predicted pixel coordinate. We define the loss function for the intrinsics constrains as:

\begin{equation}
    los{{s}_{intr}}=\sum\limits_{i=0,j=0}^{i=n,j=m}{\left\| \frac{gt\_{{c}_{ij}}}{\sqrt{{{h}^{2}}+{{w}^{2}}}}-\frac{pd\_{{c}_{ij}}}{\sqrt{{{h}^{2}}+{{w}^{2}}}} \right\|_{2}^{2}}
    \label{eq:14}
\end{equation}

where $h,w$ represent the height and width of the image, and $gt\_{{c}_{ij}}$ denotes the ground-truth coordinate of  ${c}_{ij}$, which can be detected in the images including the AprilTag cube.

Solving for the extrinsics with the condition of providing intrinsic constraints is a typical Perspective-n-Point (PnP) problem. As a typical method to solve PnP, Bundle Adjustment (BA) can be described as to optimize extrinsics minimizing the difference between the projected 3D scene points in the images and their corresponding 2D image feature points. The loss function in BA is defined as:

\begin{equation}
    los{{s}_{BA}}=\frac{1}{2}\sum\limits_{i=0,j=0}^{i=n,j=m}{\left\| \frac{gt\_{{p}_{ij}}}{\sqrt{{{h}^{2}}+{{w}^{2}}}}-\frac{pd\_{{p}_{ij}}}{\sqrt{{{h}^{2}}+{{w}^{2}}}} \right\|_{2}^{2}}
    \label{eq:15}
\end{equation}

where $h,w$ also represent the height and width of the image, and $gt\_{{p}_{ij}}$ denotes the AprilTag corners detected in $Pack1$. $pd\_{{p}_{ij}}$ represents the reprojected coordinates of the AprilTag corners during the PnP optimization.

In addition to the loss terms for camera parameters, the last loss function is the photometric loss commonly used in NeRF, defined as:

\begin{equation}
\begin{aligned} 
    los{s_{rgb}} = \sum {\left\| {\hat C(r) - C(r)} \right\|_2^2}
\end{aligned} 
\label{eq:16}
\end{equation}

where $\hat C(r)$ represents the ground truth RGB colors for ray $r$, and $C(r)$ represents the RGB colors predicted by the NeRF method.

\subsubsection{World coordinate system}

Although Eq.\eqref{eq:14} and Eq.\eqref{eq:15} differ only in their coefficients, their implications are entirely different. Eq.\eqref{eq:14} uses the auxiliary images from $Pack2$ to constrain the intrinsics and address the coupling issue, with the extrinsics' world coordinate system centered at each AprilTag cube. Since the cube in $Pack2$ is moved randomly, its position is unrelated to the multi-camera system's world coordinate system. Consequently, the extrinsics optimized from Eq.\eqref{eq:14} are redundant parameters, corresponding to the $Extrinsics$ $Local$ in Fig.\hyperref[Fig.2]{\ref{Fig.2}}, and are not involved in the NeRF training stage.

Eq.\eqref{eq:15} utilizes the auxiliary images from $Pack1$ to constrain the extrinsics of all cameras and to serve as the initial setup for these parameters. We define the coordinate system in $Pack1$ as the world coordinate system for the multi-camera system. In this scenario, the optimized extrinsics represent the extrinsics of each camera within the system and are integrated into the joint optimization with NeRF.

The origin of the world coordinate system is defined at the geometric center of the cube. The X-axis points towards the AprilTag labeled with number 1, the Y-axis points towards number 2, and the Z-axis points towards number 4.

\begin{figure*}[htbp]
\centering
	\includegraphics[width=\linewidth]{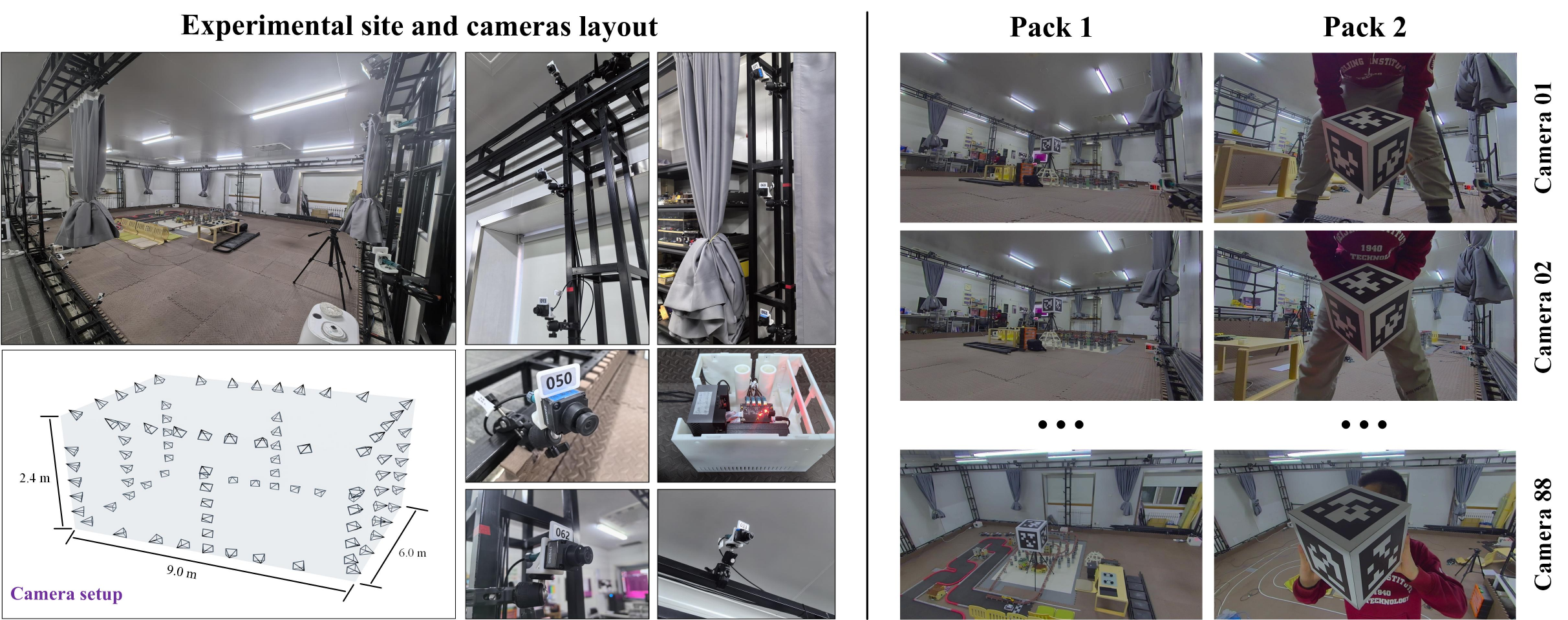}
	\caption{Multi-camera image acquisition system in real-world and captured images containing AprilTags. The left side details the dimensions of our experimental site and the camera setup, utilizing a total of 88 cameras across the scene. The right side presents the real-world scene data we collected, divided into two sets. Consistent with the descriptions in Fig.\hyperref[Fig.2]{\ref{Fig.2}}, $Pack1$ represents the Global package, while $Pack2$ represents the Local package. Because the site covered by the cameras is also the scene for NeRF representation, we use the images in $Pack1$ for training.}
    \label{Fig.7}
\end{figure*}

\subsubsection{NeRF with Bundle Adjustment}

Lin et al. first proposed Bundle-Adjusting Neural Radiance Fields (BARF), which addresses the problem of achieving NeRF training with imperfect camera extrinsics. One of the key advantages of BARF is its independence from requiring extra information. It only needs simple adjustments to the coarse-to-fine NeRF architecture to achieve the joint optimization for extrinsics and NeRF.

We attempted to extend this method by incorporating the estimation of intrinsics into the BARF framework. As a downstream component in the workflow, the rendering network is susceptible to fluctuations in upstream inputs, suggesting that training the rendering network could be ineffective when camera parameters change drastically. Moreover, the number of optimized parameters in the photometric loss branch far exceeds that of the intrinsic and extrinsic reprojection loss branches, which prolongs the training duration. Therefore, initializing camera parameters and subsequently refining them during rendering network training emerges as an efficient
approach.

The training sequence is illustrated in Fig.\hyperref[Fig.6]{\ref{Fig.6}}. During the camera parameters initialization stage, coarse camera parameters are obtained using the calibration data $Pack1$ and $Pack2$. In the global optimization stage, we employ the intrinsic reprojection loss branch to impose constraints on the principal point coordinates of the camera. We leverage the progressive alignment method from BARF to optimize the extrinsic parameters of the multi-camera acquisition system. In the fine-tuning stage, we freeze the extrinsic parameters and only refine the intrinsic parameters to obtain the final results.

\subsection{Multi-Camera System Design and Dataset}

In this section, we first introduce the multi-camera image acquisition system we constructed in the real world (Sec 3.5.1). Next, we present the proposed datasets, which includes both synthesis and real-world scenarios (Sec 3.5.2). 

\subsubsection{Multi-Camera Image Acquisition System}

The motivation for MC-NeRF comes from the multi-camera image acquisition system constructed by ourselves. The dimensions of our site are a rectangular area measuring 9.0m $\times$ 6.0m $\times$ 2.4m. There are a total of 88 different cameras, each with independent camera parameters. Based on our experience, the calibration parameters of one camera cannot be shared with others, even though these cameras are claimed to be of the same model. The cameras are mounted on the truss in a uniform distribution, aligning with the setup commonly found in multi-camera image acquisition systems, such as facial imaging and motion capture systems. Fig.\hyperref[Fig.7]{\ref{Fig.7}}. shows the detailed information of our site, as well as the scene data captured using our system.

\subsubsection{Datasets of Multi-Camera Systems}

Existing datasets, such as Synthesis, LLFF, and Mip-NeRF 360, typically capture scenes with a single camera, varying only in the extrinsics, which can not reflect the characteristics of multi-camera image acquisition systems. While some datasets for human reconstruction, such as HumanRF \cite{ref63} and DNA-Rendering \cite{ref64}, use multi-camera systems, they fail to provide the auxiliary images. Our goal is to lower the barriers to using the NeRF method, and most current datasets are primarily designed to enhance rendering quality, which has motivated us to develop our own dataset.

\begin{figure*}[htbp]
\centering
	\includegraphics[width=\linewidth]{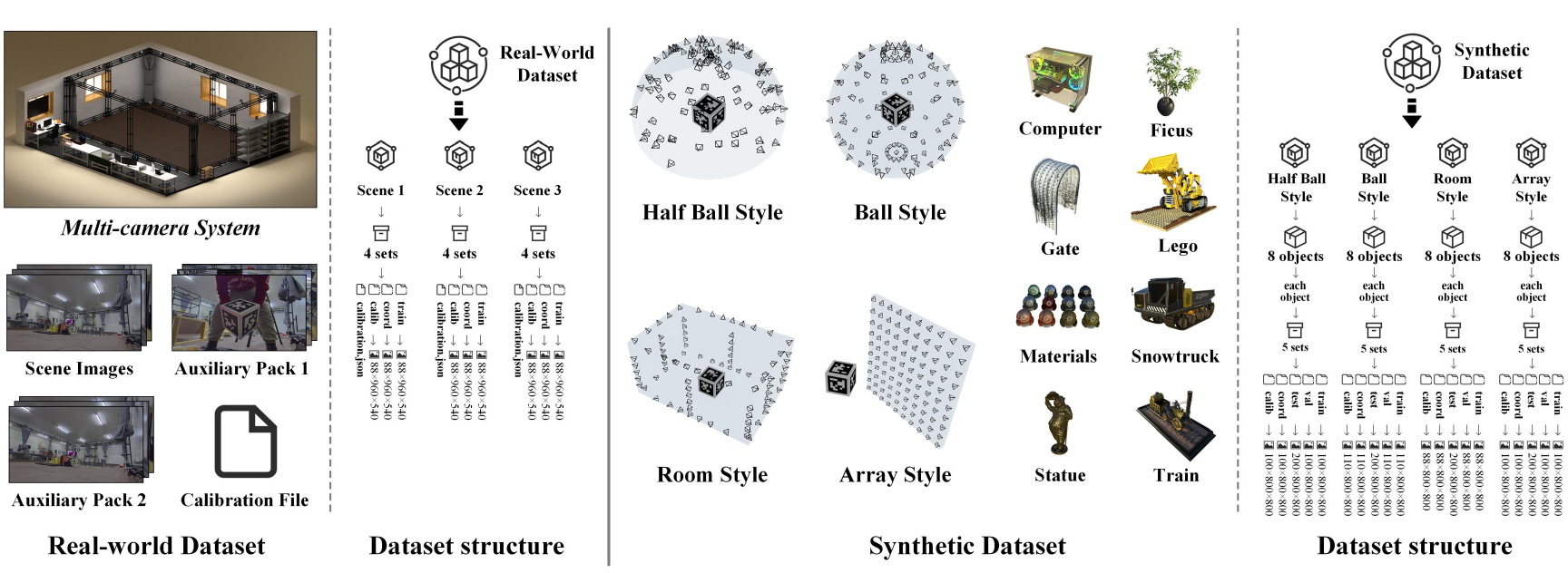}
	\caption{Dataset Composition and Structure: The proposed datasets are divided into two parts: the left side illustrates the content and structure of the real-world dataset, while the right side displays the corresponding content and structure of the synthetic dataset. The real-world dataset comprises three scenes, each containing four components: the images of the scene to be reconstructed, two sets of auxiliary images, and calibration files for all cameras. These calibration files are critical for evaluating the accuracy of the estimated camera parameters. The synthetic dataset consists of four unique styles, each designed to simulate a different multi-camera acquisition system, with eight objects per style. Unlike the real-world dataset, the synthetic dataset additionally includes dedicated validation and test data.}
	\label{Fig.8}
\end{figure*}

Our public datasets consist of two parts: a synthetic dataset and a real-world scene dataset. For the synthetic dataset, we designed four packages based on commonly used multi-camera image acquisition systems. The camera distributions in these datasets include half-ball style (as used by the original NeRF), ball style, room style, and array style. The datasets contain a total of 32 groups, with 8 different objects for each style. The different sampling styles involve varying numbers of cameras, resulting in different numbers of training images. 

The synthetic datasets are generated using Blender 3.3.3, and the intrinsic parameters for each camera within each style are different. For example, in the HalfBall Style acquisition system, there are a total of 100 cameras with varying intrinsic parameters. The training data is collected using these cameras according to their depicted positions. Similarly, the validation data is also collected using these cameras, but each camera's position differs from that of the training set. For each style, the number of validation images is equal to the number of training images. The resolution of each image is $800 \times 800$. 

For the real-world scene dataset, we collect three sets of images using our custom-built system. Each scene includes 88 cameras with different parameters, and calibrated camera parameters are provided as ground truth for each camera. The resolution of each image is $960 \times 540$. Both synthetic and real-world datasets include two sets of auxiliary images to validate our method. The details of the datasets are illustrated on Fig.\hyperref[Fig.8]{\ref{Fig.8}}

It's worth noting that due to the images being collected from cameras with different intrinsic parameters, determining which camera to use for generating the test data is challenging. To showcase the performance of the proposed method across various camera intrinsic parameters, we generate the 200 test images in each group by continuously varying the intrinsic parameters along the camera's motion trajectory for synthetic dataset. Detailed sampling procedures and parameter variations have been provided on the project website.

\section{EXPERIMENTS}

All of our experiments were conducted on an NVIDIA RTX 3090 with 24GB of memory, using PyTorch version 1.12.1. For all comparative experiments, we utilized the publicly available source code from their respective papers. To facilitate the replication of our experiments, we did not adjust the hyperparameters of the open-source code, except for modifying the data loading workflow. The goal of the experiments is to answer the following questions:

\begin{enumerate}

\item How does the coupling issue affect the joint optimization process? In most methods that jointly optimize camera parameters and NeRF, 2D image alignment experiment is an effective way to validate the performance of joint optimization. How does the proposed method perform in the 2D image alignment experiments?

\item Sec 3.3.1 highlights that if the selected 3D points lie on the same plane, a degenerate case could occur. Does the proposed image acquisition scheme avoid this issue?

\item The author claims that $Pack1$ can be used to constrain the extrinsics and provide initial extrinsics. However, theoretically, $Pack2$ is already capable of providing constraints for solving all parameters. Is $Pack1$ necessary? Can the existing methods achieve accurate results if the extrinsics are initialized with random values?

\item How does the performance of the proposed method compare to other joint optimization approaches? Existing methods typically rely on straightforward combined strategies, such as using off-the-shelf SFM libraries to provide initial camera parameters. How does the performance of the proposed method compare to this strategy?

\end{enumerate}

\begin{figure*}[htbp]
\centering
\includegraphics[width=\linewidth]{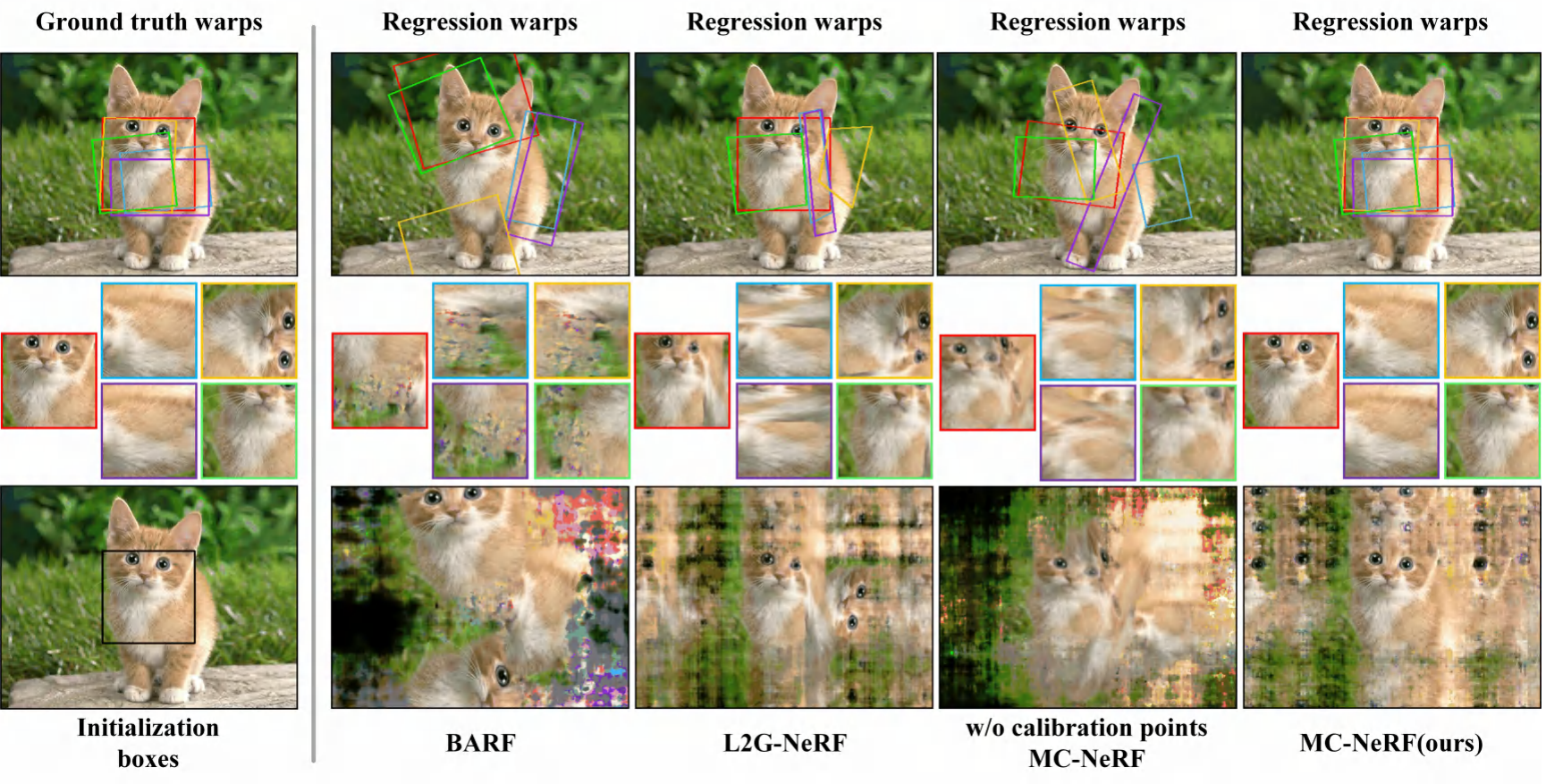}
\caption{The performance of the 2D image alignment experiment. We visualize the optimized warps (top row), the patch reconstructions in corresponding colors (middle row), and recovered image representation from ${f_{mlp}}$ (bottom row). MC-NeRF is able to recover accurate alignment and high-fidelity image reconstruction. This indicates that joint optimization of intrinsics, extrinsics and NeRF involves more than merely converting these parameters into learnable variables and applying backpropagation.}

\label{Fig.9}
\end{figure*}

\subsection{Exp.1 2D Alignment for Coupling Issue}

2D neural image alignment problem can be described as: let $x$ be the 2D pixel coordinates, we aim to optimize a 2D neural field parameterized as the weights  of a multilayer perceptron (MLP) ${f_{mlp}}$:

\begin{equation}
\begin{aligned} 
  RGB(x) = {f_{mlp}}(Tx;w)
\end{aligned} 
\label{eq:17}
\end{equation}
while also solving for geometric transformation as $T = [R{\rm{|}}T]$, and $[R{\rm{|}}T]$ is the rigid rotation and translation in 2D space. 

Compared to the previously introduced 2D image alignment problem, we have increased its complexity even further, resulting in the updated problem formulation:

\begin{equation}    
\begin{aligned} 
  RGB(x) = {f_{mlp}}(TLx;w)
\end{aligned} 
\label{eq:18}
\end{equation}

Where $L$ represents a linear transformation, simulating the computation flow of projection from pixel coordinates to camera coordinates. The form of $L$ is consistent with that of the intrinsic matrix $K$ and is defined as follows:

\begin{table}[htbp]
\caption{Quantitative results of neural image alignment experiment. MC-NeRF optimizes for high-quality alignment and patch reconstruction.}
    \centering
    \renewcommand\arraystretch{1.3}
    \resizebox{0.48\textwidth}{!}{
    \begin{tabular}{ccccc}
    \toprule
       & Corner Error$\downarrow$ & Patch & Patch & Patch\\ 
       & (pixels) & PSNR$\uparrow$ & SSIM$\uparrow$ & LPIPS$\downarrow$ \\
    \hline
    BARF & 85.6157 & 14.2270 & 0.5149 & 0.6641 \\
    \hline
    L2G-NeRF & 55.3545 & 20.5440 & 0.7884 & 0.3364\\
    \hline
    MC-NeRF(w/o) & 55.3096 & 15.9662 & 0.5923 & 0.6268\\
    \hline
    MC-NeRF(ours) & \cellcolor{red!20}0.4084 & \cellcolor{red!20}40.8309 & \cellcolor{red!20}0.9828 & \cellcolor{red!20}0.0318\\
    \bottomrule
    \end{tabular}}
    \label{Table 1}
\end{table}

\begin{equation}    
L = \left[ {\begin{array}{*{20}{c}}
{{a_1}}&0&{{b_1}}\\
0&{{a_2}}&{{b_2}}\\
0&0&1
\end{array}} \right] 
\label{eq:19}
\end{equation}

$a_1$, $a_2$, $b_1$, $b_2$ are four parameters to be optimized, used to simulate inaccurate $f_x$, $f_y$, $u_0$, $v_0$ in intrinsics. It's important to note that the term $T$ in Eq.\eqref{eq:17} is defined as a homography matrix in original experiment. However, since the transformation requirements for 3D space in the NeRF series methods involve only rotation and translation, we redefine $T$ as $[R{\rm{|}}T]$ instead. 

We choose the same image "cat" as BARF used from ImageNet \cite{ref51}. We select five patches sampled from the original image, apply linear transformations and rigid perturbations, and optimize Eq.\eqref{eq:18} to find the transformation $L$ and $T$ for each patch. Simultaneously, we train the neural network ${f_{mlp}}$ to represent the entire image. In all test models, a 6-layer fully connected network with 256 neurons each is respectively defined. The training epochs are set to 50,000 to ensure convergence for all methods. The training process is demonstrated on our project webpage. 

\begin{figure}[htbp]
\centering
\includegraphics[width=\linewidth]{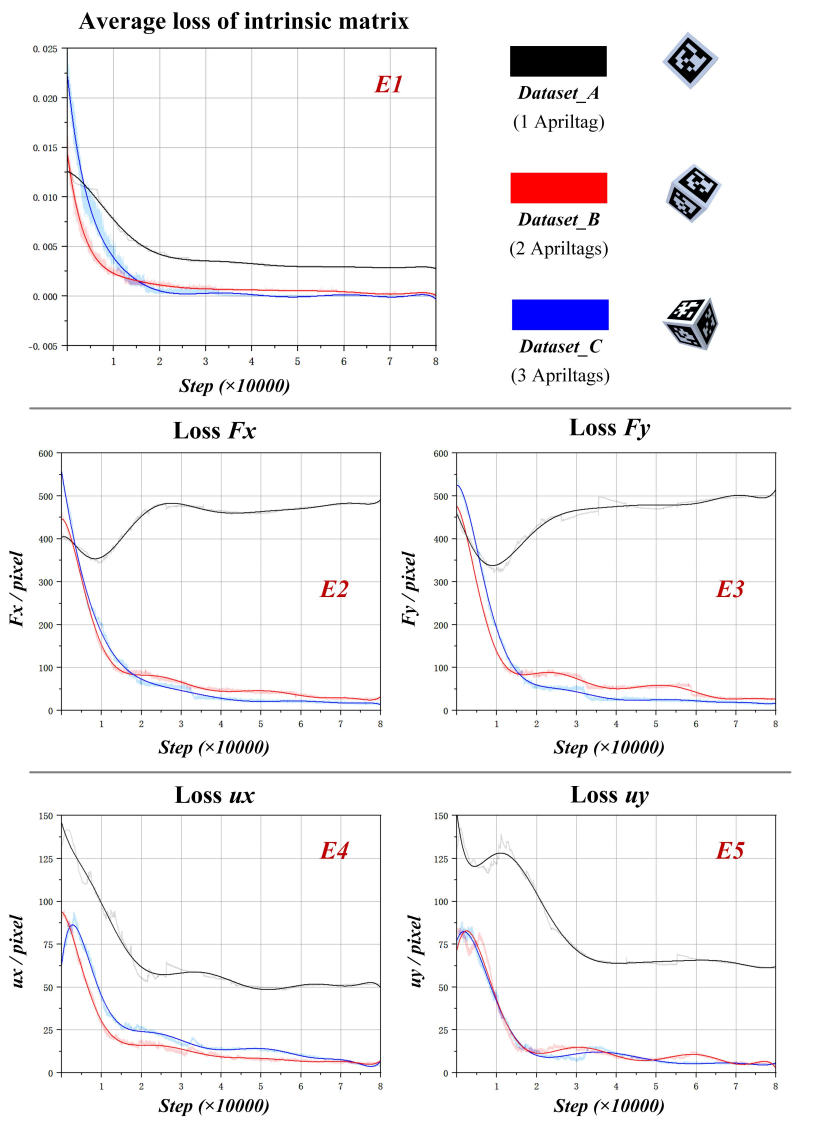}
\caption{Comparison results of estimated intrinsics with different numbers of AprilTags. With at least two AprilTags included in the auxiliary image, we can achieve the expected results. However, when including only one AprilTag, degeneration case is showed up, leading to poor results. Specifically, the black curve in $E1$ fails to converge to zero, while the blue and red curves both converge near zero. This means that when the selected 3D points are distributed on the same plane, degeneration occur, which is consistent with our conclusion.}
\label{Fig.10}
\end{figure}

The joint optimization of additional intrinsics can be equivalent to adding an additional linear transformation $L$ to the original rigid transformation. To decouple this linear transformation from rigid transformation, we randomly selected six pairs of calibration points as constraints for each patch. This constraint is analogous to the correspondences obtained by AprilTag cube in 3D space. 

The render results are shown in Fig.\hyperref[Fig.9]{\ref{Fig.9}}. The inclusion of linear transformations leads to both BARF and L2G-NeRF being unable to produce accurate results, which indicates that the coupling issue indeed exists. Even in cases where matches are not provided, proposed method also faces challenges in achieving precision results. With the condition of using calibration points for constraints, the accurately estimation of $L$ and $T$ can be obtained.

\begin{table}[htbp]
\caption{Quantitative results of \textit{\textbf{Exp.2}}. The loss items on the left represents the errors when estimating intrinsics using images containing different numbers of AprilTags. It can be observed that when using only one AprilTag for estimation (Dataset A), accurate intrinsics cannot be obtained, leading to a degenerate case.}
    \centering
    \renewcommand\arraystretch{1.3}
    \resizebox{0.48\textwidth}{!}{
    \begin{tabular}{ccccccc}
    \toprule
      & Train Step & $10000$ & $20000$ & $40000$ & $60000$ & $80000$
    \\\hline
    \multirow{3}{*}{$Loss\_Fx$} & Dataset A & 370.06 & 458.82 & 460.30 & 474.59 & 487.35 \\
     &Dataset B&153.53&88.98&53.87&32.47&26.64 \\
     &Dataset C&198.08&69.21&26.73&19.41&17.41 \\
    \hline
    \multirow{3}{*}{$Loss\_Fy$} & Dataset A  & 343.69 & 429.86 & 462.75 & 470.78 & 511.29 \\
     &Dataset B&144.23&92.43&55.34&32.47&26.64 \\
     &Dataset C&197.32&57.01&25.65&21.34&18.91 \\
     \hline
    \multirow{3}{*}{$Loss\_Ux$} & Dataset A  & 99.42 & 62.91 & 58.71 & 51.43 &51.61 \\
     &Dataset B&23.17&15.83&12.85&9.27&7.75 \\
     &Dataset C&30.73&19.57&11.37&9.77&7.10 \\
     \hline
    \multirow{3}{*}{$Loss\_Uy$} & Dataset A  & 129.51 & 105.26 & 64.49 & 68.22 & 61.94 \\
     &Dataset B&44.52&14.98&13.056&12.48&9.73 \\
     &Dataset C&43.17&13.65&14.21&8.31&7.19 \\
    \bottomrule
    \end{tabular}}
    \label{Table 2}
\end{table}

The quantitative results are shown in Table \ref{Table 1}. $Corner Error$ measures the absolute difference between the predicted and ground truth coordinates. This metric is calculated for the four vertices of a patch after applying the optimized $L$ and $T$. PSNR, SSIM, and LPIPS represent the rendering quality metrics for each patch. It can be seen that MC-NeRF achieved the best performance after decoupling constraints were applied.

\subsection{Exp.2 Capture Scheme for Degeneration Case}

In this experiment, we aim to verify whether the proposed auxiliary image capture scheme can avoid the degeneration case. We conduct experiments using AprilTag images collected by 70 cameras with different FOVs. The dataset consists of three groups: the images in the first group contain only one AprilTag in each image, denoted as $Dataset\_A$.  This group ensures that the selected 3D points lie in the same plane. The images in the second group contain two AprilTags, denoted as $Dataset\_B$. According to the conclusion in Sec 3.3.1, the results trained by this dataset can successfully obtain the intrinsics and avoid degeneration case. The images in the third group contain three AprilTags, denoted as $Dataset\_C$. We believe that the results trained with this group will have better accuracy. Here, $Fx$ and $Fy$ are the focal lengths of the camera in the horizontal and vertical directions, while $ux$ and $uy$ are the coordinates of the principal point.

The results are presented in Fig. \hyperref[Fig.10]{\ref{Fig.10}} and Table \ref{Table 2}. In Fig. \hyperref[Fig.10]{\ref{Fig.10}}, average loss of intrinsics refers to the average absolute error between predicted intrinsics and ground truth. In this scenario, the intrinsics cannot be obtained with the images contain only 1 AprilTag ($Dataset\_A$). This case exhibits performance to the degeneration case. Specifically, as illustrated by the black curves in $E2$ and $E3$, $Fx$ and $Fy$ fail to converge, and the losses in $E4$ and $E5$ exhibit significant deviations from $0$. Moreover, Table \ref{Table 2} reveals that the accuracy of the results is significantly inferior to the other two cases.

When the images we used contain more than two AprilTags ($Dataset\_B$, $Dataset\_C$), the estimation can be guaranteed to converge. It can be observed from the quantified data in the table that the parameters trained using the dataset containing three AprilTags achieve higher accuracy, as demonstrated in $Fx$, $Fy$, $ux$, and $uy$. In comparison, the parameters trained with the dataset containing 2 AprilTags converge more rapidly.

In summary, the image capture scheme we proposed is effective and can prevent the degeneration case. If the provided 3D points lie on the same plane, the degeneration case will occur. Including more than two AprilTags in a single auxiliary image can provide effective intrinsic constraints, ensuring that the joint optimization process yields accurate results.

\subsection{Exp.3 Initialization of Extrinsics Estimation}

\begin{figure*}[htbp]
\centering
\includegraphics[width=\linewidth]{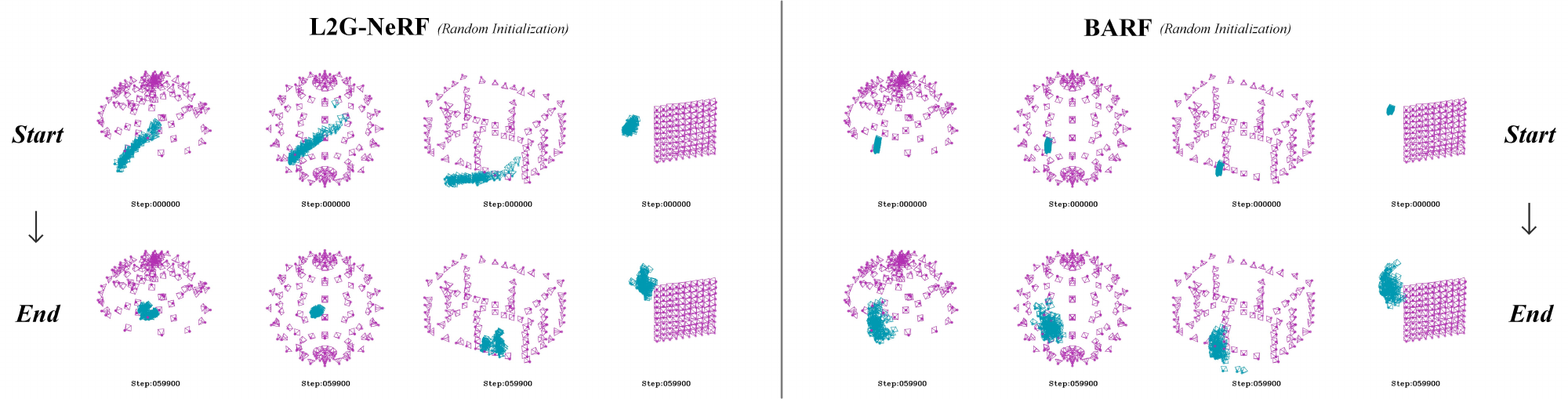}1
\caption{Comparison of extrinsics estimation performance between BARF and L2G-NeRF with the random initialization values.}
\label{Fig.11}
\end{figure*}

In this experiment, we aim to verify the necessity of extrinsics estimation with $Pack1$. Both BARF and L2G-NeRF address the joint optimization of NeRF and extrinsics by adding perturbations to the ground truth extrinsics. This leads to camera pose drift, which the methods then attempt to rectify during the joint optimization process. 

However, the initial extrinsics of each camera in practical applications still requires huge preparation work, especially for the multi-camera acquisition systems with a large number of  cameras. Therefore, in this experiment we randomly provide the initial extrinsics of all cameras and attempt to perform extrinsics estimation on four styles of scenes using the above mentioned two methods. This is done to assess the performance of the previous methods under the condition where no potential poses information is provided.

The results are presented in Fig. \hyperref[Fig.11]{\ref{Fig.11}}. The first row illustrates the randomly initialized extrinsics of the two methods across the four scenarios, while the second row depicts the stabilized extrinsics after training. When the extrinsics are initialized randomly, both methods struggle to acquire accurate values. As our proposed method employs the progressive alignment technique introduced by BARF in the joint optimization, it becomes evident that an effective initialization of the extrinsics is essential. Further details regarding the convergence process are provided in our supplementary project webpage.

\begin{table}[htbp]
\caption{Comparison of extrinsics estimation performance between BARF and L2G-NeRF with the initialization extrinsics obtained from AprilTag cube. $Inherited$ $Value$ represents the initial camera position obtained directly using the reprojection error method.}

\centering
    \renewcommand\arraystretch{1.3}
    \resizebox{0.48\textwidth}{!}{
    \begin{tabular}{ccccc}
    \toprule
       Style&  & Inherited Value & L2G-NeRF & BARF 
    \\\hline
    \multirow{2}{*}{Ball} & $Loss\_R$ & 0.0392 & 0.0054 & 0.0050 \\
     &$Loss\_T$ &0.1355&0.0463&0.0509 \\
    \hline
    \multirow{2}{*}{HalfBall} & $Loss\_R$  & 0.0601 & 0.0183 & 0.0114 \\
     &$Loss\_T$&0.1891&0.0224&0.0321\\
     \hline
    \multirow{2}{*}{Room} & $Loss\_R$  & 0.0473 & 0.0048 & 0.0027 \\
     &$Loss\_T$&0.1844&0.0366&0.0284\\
     \hline
    \multirow{2}{*}{Array} & $Loss\_R$& 0.0561 & 0.0727 & 0.0915\\
     &$Loss\_T$&0.2154&0.0149&0.0287\\
    \bottomrule
    \end{tabular}}
    \label{Table 3}
\end{table}

\begin{figure*}[htbp]
\centering
\includegraphics[width=\linewidth]{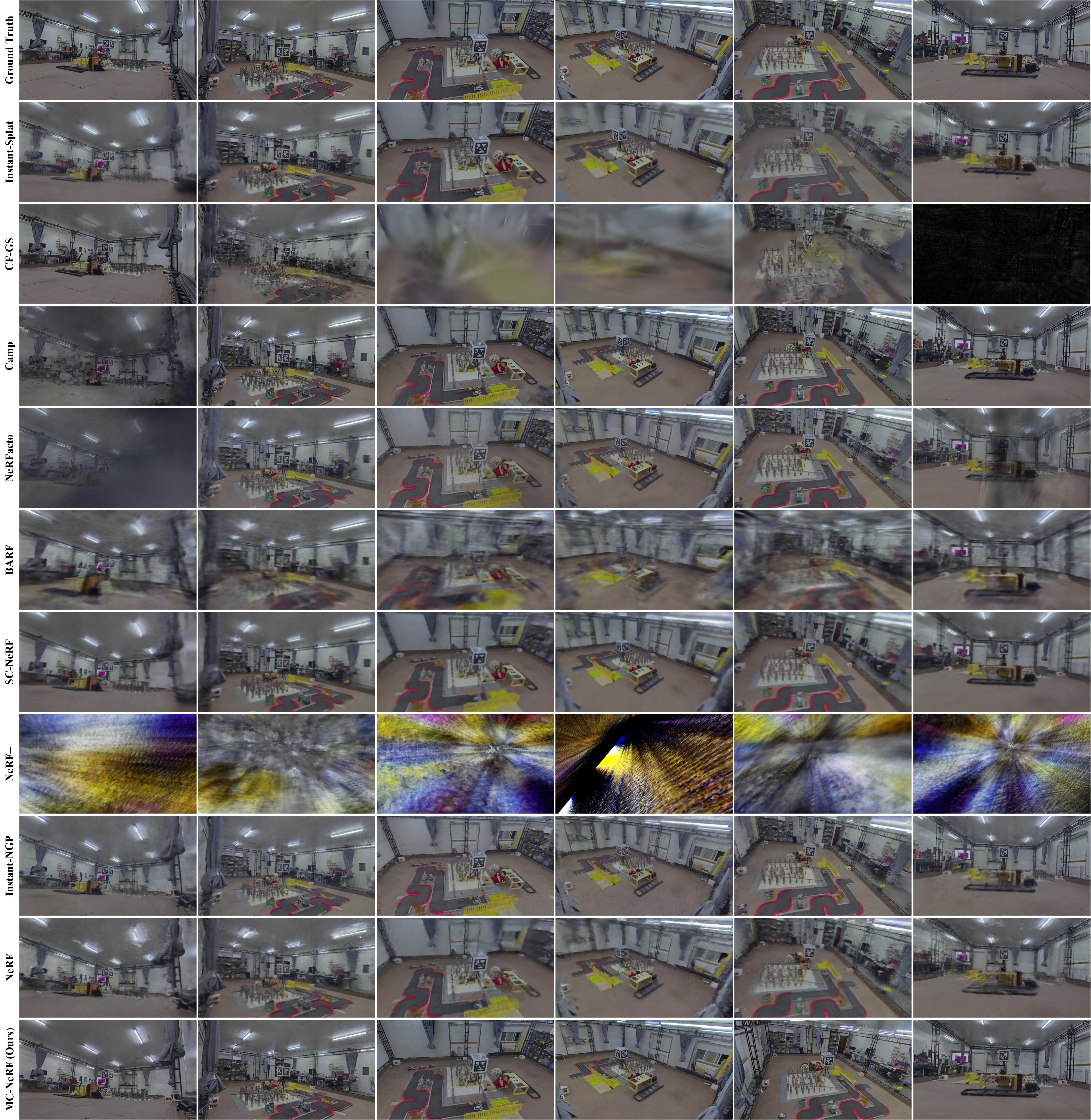}
\caption{Rendering results of different methods in real-world dataset. We select the $Lab1$ scene from proposed real-world dataset to showcase rendering quality, while the results for the remaining two scenes are provided in the supplementary materials and on project website. In our comparative experiments, InstantSplat and CF Gaussian autonomously estimate camera parameters without any provided initializations. Methods such as CAMP, NeRFacto, and BARF, which are initialized using camera parameters provided by Colmap, are limited to optimizing only the extrinsic parameters. In contrast, SC-NeRF and NeRF$--$, also initialized with Colmap, are capable of optimizing both intrinsic and extrinsic parameters. Instant-NGP, which lacks camera parameter optimization capabilities, serves as a baseline reference. MC-NeRF utilizes auxiliary images for initialization, and it is also capable of optimizing both intrinsic and extrinsic parameters. The results demonstrate that our method achieved the competitive rendering performance among all compared methods.}
\label{Fig.12}
\end{figure*}

Random initialization of extrinsics can lead to optimization failure. We provide auxiliary images from $Pack1$ to assist in obtaining initial values of the extrinsics. With these initial values provided, the performance of the above two methods is shown in Table \ref{Table 3}. Inherited Value refers to the initial values provided by AprilTag cube, $Loss\_R$ and $Loss\_T$ denote the losses associated with the ground truth rotation and translation data. These losses are defined as follows:

\begin{equation}
    \begin{aligned}
        \left\{\begin{matrix}
Loss\_R=\sum_{i=0}^{n}mean(||\hat{R}_{i}-R_{i}||_{2}^{2})/n\\ 
Loss\_T=\sum_{i=0}^{n}mean(||\hat{T}_{i}-T_{i}||_{2}^{2})/n
\end{matrix}\right.
    \end{aligned}
\end{equation}

${\hat R_i}$ and ${R_i}$  represent the ground truth and predicted rotation matrices respectively, while ${\hat T_i}$ and ${T_i}$ represent the ground truth and predicted translation matrices. $mean()$ represents the mean function. $n$ represents the total number of poses, which is equal to the number of images in the dataset. It can be observed that when using the initial intrinsics, these methods achieved high performance under all four styles.

In summary, it is necessary to use the auxiliary image set $Pack1$ to provide initial extrinsics for the camera. Randomly initializing extrinsics cannot guarantee obtaining accurate results. By using AprilTag cube to obtain initial values, we can subsequently achieve precise extrinsics results.

\subsection{Exp.4 Comparison with Other methods}

In this experiment, we aim to validate the effectiveness of our method in real-world scenarios. We selecte NeRF series methods, including CAMP, NeRF$--$, BARF, SC-NeRF, Nerfacto and Instant-NGP as baselines. Additionally, we included 3D Gaussian-based methods, specifically CF-3D Gaussian and InstantSplat, as additional baselines. For methods that require initial camera parameters, we rely on the widely used COLMAP for initialization. In contrast, our approach incorporates auxiliary images to assist in this process. The real-world dataset consists of three scenes: $Lab1$, $Lab2$, and $Lab3$. Due to space limitations, we present the training results of $Lab1$ in the main manuscript, while the rendered results for the other two scenes are provided in the supplementary materials.

The experimental results are presented in Fig.\hyperref[Fig.12]{\ref{Fig.12}}, Table \ref{Table 4}, and Table \ref{Table 5}. Regarding image rendering quality, in Fig.\hyperref[Fig.12]{\ref{Fig.12}}, we select six views generated by different methods randomly, and the results demonstrate that our method achieves competitive performance. In this experiment, INGP is used as the baseline without any camera parameter optimization. Only SC-NeRF, NeRF$--$, and our method jointly optimize both the intrinsic and extrinsic parameters, while the remaining methods optimize only the extrinsic parameters. It is important to note that NeRF$--$ is the only method that failed to render in all cases. In the original paper, the authors explicitly state that their approach is only applicable to forward-facing scenes, which is consistent with our experimental results.
\begin{table*}[htbp]
\caption{Quantitative camera parameters estimation results of different methods on real-world datasets. The table shows that our method achieves competitive results in all three aspects: camera intrinsics, camera extrinsics, and scene reconstruction scale.}
\centering
    \renewcommand\arraystretch{1.3}
    \resizebox{\textwidth}{!}{
    \begin{tabular}{c|c|ccccccc}
    \toprule
       Scene& Method & $Loss\_Fx \downarrow$ & $Loss\_Fy\downarrow$ & $Loss\_Cx\downarrow$ & $Loss\_Cx\downarrow$ & $Loss\_R\downarrow$ & $Loss\_T\downarrow$ & $Loss\_Scale\uparrow$ 
    \\\hline
    \multirow{9}{*}{Lab 1} & InstantSplat & 20.2304 & 20.2586 & 8.3652 & 8.7434 & 0.2619 & 0.0982 & 0.0478 \\
     & CF Gaussian & 79.1077 & 79.1313 & 7.5821 & 8.4547 & 0.4124 & 1.8820 & 0.7239 \\
     & CAMP & 15.7198 & 15.5267 & 7.3414 & 8.5711 & 0.0258 & 0.0182 & 0.2195 \\
     & NeRFacto & 7.3588 & 7.1762 & 11.3652 & 11.7434 & 0.2679 & 0.0588 & 0.7946 \\
     & BARF & 7.3588 & 7.1762 & 11.3652 & 11.7434 & 0.2661 & 0.8986 & 0.2596 \\
     & SC-NeRF & 9.3722 & 13.0315 & 4.8873 & 9.3619 & 0.0135 & 0.0373 & 0.7991 \\
     & NeRF$--$ & 163.1592 & 165.1681 & 18.3652 & 18.7434 & 0.4554 & 0.0177 & 0.0839 \\
     & NeRF & 7.3588 & 7.1762 & 11.3652 & 11.7434  & 0.2721 & 0.0749 & 0.3839 \\
     & Instant-NGP & 7.3588 & 7.1762 & 11.3652 & 11.7434 & 0.2721 & 0.0749 & 0.3839 \\
     & MC-NeRF (ours) & \cellcolor{red!20} 4.1757 & \cellcolor{red!20}4.1002 & \cellcolor{red!20}3.8026 & \cellcolor{red!20}4.2973 & \cellcolor{red!20}0.0011 & \cellcolor{red!20}0.0129 & \cellcolor{red!20}0.9625 \\
    \hline
    \multirow{9}{*}{Lab 2} & InstantSplat & 19.4612 & 19.4894 & 8.3652 & 8.7434 & 0.2524 & 0.2459 & 0.0504 \\
     & CF Gaussian & 79.1077 & 79.1313 & 7.5821 & 8.4547 & 0.4328 & 2.0865 & 0.7155 \\
     & CAMP & 16.8892 & 16.8645 & 7.3414 & 8.5711 & 0.0089 & 0.2135 & 0.2224 \\
     & NeRFacto & 8.9282 & 8.4323 & 12.3652 & 12.7434 & 0.2381 & 0.2415 & 0.8015 \\
     & BARF & 8.9282 & 8.4323 & 12.3652 & 12.7434 & 0.2384 & 1.0924 & 0.2912 \\
     & SC-NeRF & 8.5852 & 15.4368 & 4.9523 & 9.1950 & 0.0165 & 0.2212 & 0.7947 \\
     & NeRF$--$ & 176.5705 & 160.6003 & 18.3652 & 18.7434 & 0.4405 & 0.1968 & 0.0805 \\
     & NeRF & 8.9282 & 8.4323 & 12.3652 & 12.7434  & 0.2424 & 0.5393 & 0.3902 \\
     & Instant-NGP & 8.9282 & 8.4323 & 12.3652 & 12.7434  & 0.2424 & 0.5393 & 0.3902 \\
     & MC-NeRF (ours) & \cellcolor{red!20}3.9113 & \cellcolor{red!20}3.8856 & \cellcolor{red!20}4.1121 & \cellcolor{red!20}4.0004 & \cellcolor{red!20}0.0067 & \cellcolor{red!20}0.0286 & \cellcolor{red!20}0.9754 \\
     \hline
    \multirow{9}{*}{Lab 3} & InstantSplat & 19.2140 & 19.2423 & 8.3652 & 8.7434 & 0.2881 & 0.0557 & 0.0477 \\
     & CF Gaussian & 79.1077 & 79.1313 & 7.5821 & 8.4547 & 0.4016 & 1.9288 & 0.7345 \\
     & CAMP & 19.2211 & 19.3228 & 7.3414 & 8.5711 & 0.0028 & 0.0435 & 0.2173 \\
     & NeRFacto & 7.3223 & 7.3193 & 12.8605 & 12.5431 & 0.2466 & 0.0720 & 0.7861 \\
     & BARF & 7.3223 & 7.3193 & 12.8605 & 12.5431 & 0.2453 & 0.8915 & 0.2552 \\
     & SC-NeRF & 11.9082 & 15.5652 & \cellcolor{red!20}4.5782 & 9.3252 & 0.0159 & 0.0561 & 0.7914 \\
     & NeRF$--$ & 129.3443 & 110.5904 & 10.6957 & 12.1557 & 0.4553 & \cellcolor{red!20}0.0102 & 0.1024 \\
     & NeRF & 7.3223 & 7.3193 & 12.8605 & 12.5431  & 0.2495 & 0.1509 & 0.3813 \\
     & Instant-NGP & 7.3223 & 7.3193 & 12.8605 & 12.5431  & 0.2495 & 0.1509 & 0.3813 \\
     & MC-NeRF (ours) & \cellcolor{red!20}4.2264 & \cellcolor{red!20}4.1906 & 4.9588 & \cellcolor{red!20}4.0681 & \cellcolor{red!20}0.0015 & 0.0341 & \cellcolor{red!20}0.9755 \\
    \bottomrule
    \end{tabular}}
    \label{Table 4}
\end{table*}
Table \ref{Table 4} presents the errors between the estimated camera parameters and the ground truth after joint optimization. Specifically, $loss\_Fx$ and $loss\_Fy$ represent the focal length errors, while $Loss\_Cx$ and $Loss\_Cy$ denote the principal point errors. $Loss\_R$ and $Loss\_T$ correspond to the rotational and translational errors of the extrinsic camera parameters. $Loss\_Scale$ represents the scale loss between the reconstructed scene and the ground truth scene, which is defined as follows:

\begin{equation}
    \begin{aligned}
    Loss\_Scale = \frac{{max(dist(T_p^i - T_p^j))}}{{max(dist(T_g^i - T_g^j))}}
    \end{aligned}
\end{equation}

Where $T_g$ represents the translation vector of the ground truth camera pose, and $T_p$ denotes the translation vector of the estimated camera pose. $dist$ indicates the Euclidean distance, and $i$,$j$ representing the camera indices. $Loss\_Scale$ reflects the scale ratio between the reconstructed scene and the real-world scene. Unlike other methods that rely on equivalent camera models, MC-NeRF leverages auxiliary images to incorporate real-world scale, enabling the reconstruction of a 1:1 scale model of the scene. Additionally, our method achieves more accurate intrinsic camera parameters by using constraints from the auxiliary images, effectively constraining the solution space and allowing for the joint optimization of each camera’s corresponding parameters. As shown in the table, our approach outperforms others in terms of intrinsics and extrinsics estimation, as well as scale accuracy.

Table \ref{Table 5} shows the rendering performance metrics for each scene. Each scene is equipped with 88 cameras, from which we randomly selected 80 for the training dataset, while the remaining 8 were used for the test dataset. The metrics provided are calculated based on the rendering outcomes from these 8 test datasets. As demonstrated in the table, our method achieves the best performance in terms of rendering quality. Compared to other joint optimization methods, our work primarily focuses on reducing the barriers to implementing 3D reconstruction in multi-camera systems, rather than enhancing rendering quality. In MC-NeRF, the rendering model remains the classic NeRF framework. We believe the performance of our method benefits more from accurate camera poses than from the NeRF technique itself, which differs significantly from methods such as CAMP and the 3DGS series methods.

\begin{table*}[htbp]
    \caption{Quantitative rendering results of different methods on real-world datasets. Compared to other methods, our approach also delivers competitive results in rendering quality. Unlike other works, our method retains the classic NeRF architecture, and we believe the improvement in rendering quality is driven by more accurate camera parameters rather than modifications to the NeRF model itself.}
    \centering
    \renewcommand\arraystretch{1.3}
        \resizebox{\textwidth}{!}{
    \begin{tabular}{c|ccc|ccc|ccc}
    \toprule
    \multirow{2}{*}{Scene} & \multicolumn{3}{c|}{Lab 1} & \multicolumn{3}{c|}{Lab 2} & \multicolumn{3}{c}{Lab 3}\\
    & $PSNR\uparrow$ & $SSIM\uparrow$ & $LPIPS\downarrow$ & $PSNR\uparrow$ & $SSIM\uparrow$ & $LPIPS\downarrow$ & $PSNR\uparrow$ & $SSIM\uparrow$ & $LPIPS\downarrow$ \\
    \hline
    InstantSplat & 20.3433 & 0.6086 & 0.3832 & 22.9553 & \cellcolor{red!20}0.7397 & 0.4981 & 20.5839 & 0.6233 & 0.4546 \\
    CF Gaussian & 15.7241 & 0.4878 & 0.5872 & 16.5323 & 0.5314 & 0.5491 & 15.4723 & 0.4761 & 0.6001 \\
    CAMP & 19.4850 & 0.5926 & 0.3714 & 20.7756 & 0.6816 & 0.3507 & 19.6813 & 0.5967 & 0.3654 \\
    NeRFacto & 21.3255 & 0.6801 & 0.3963 & 21.6011 & 0.7304 & 0.3517 & 22.3201 & \cellcolor{red!20}0.7002 & \cellcolor{red!20}0.3489 \\
    BARF & 18.4214 & 0.5645 & 0.7460 & 20.0294 & 0.6585 & 0.6673 & 20.8126 & 0.6342 & 0.5901 \\
    SC-NeRF & 22.3175 & 0.6555 & 0.5272 & 23.8036 & 0.7369 & 0.4795 & 22.6125 & 0.6649 & 0.5288 \\
    NeRF$--$ & 13.4507 & 0.3778 & 0.8845 & 10.8519 & 0.4552 & 0.8529 & 14.0315 & 0.4027 & 0.8561 \\
    NeRF & 22.3175 & 0.6555 & 0.5272 & 23.8036 & 0.7369 & 0.4795 & 24.0208 & 0.6935 & 0.4542 \\
    Instant-NGP & 21.0498 & 0.6097 & 0.4174 & 23.1911 & 0.7202 & 0.3428 & 20.9077 & 0.6087 & 0.4113 \\
    MC-NeRF(ours) & \cellcolor{red!20}23.1187 & \cellcolor{red!20}0.7199 & \cellcolor{red!20}0.3265 & \cellcolor{red!20}23.9901 & 0.7012 & \cellcolor{red!20}0.3277 & \cellcolor{red!20}24.3339 & 0.6923 & 0.3589 \\
    \bottomrule
    \end{tabular}}
    \label{Table 5}
\end{table*}

\section{CONCLUSIONS}
In this work, we present Multi-Camera Neural Radiance Fields (MC-NeRF), a 3D representation method designed for multi-camera image acquisition systems, which can joint optimize both camera parameters and NeRF. MC-NeRF breaks the assumption of a unique camera in previous studies and avoids providing initialized camera parameters. We analyzed the challenges of joint optimization, including coupling issue and intrinsics degenerated case. Based on this analysis, we design the framework with training sequence, calibration objects, and a calibration data acquisition strategy. Furthermore, we also provide a datasets with each image corresponding to different camera parameters, including both virtual and real-world scenes. The proposed method primarily tackles the substantial workload associated with acquiring multi-camera intrinsic and extrinsic parameters when employing multi-camera acquisition systems for 3D scene representation. Experiments demonstrate that proposed method achieves better rendering results and more accurate camera parameters. Code and models will be made available to the research community to facilitate reproducible research.

However, our method has several limitations: First, the multi-camera system requires a common field of view. Second, the images captured by these cameras must not contain image distortions. Third, in large-scale environments, $Pack1$ necessitates the use of large AprilTag cube to ensure the accuracy of parameter estimation. Based on our experience, in a multi-camera system covering an area of approximately 9.0m$\times$6.0m$\times$2.4m, using calibration blocks with a side length of 0.5m yields satisfactory reconstruction results. Employing smaller blocks significantly diminishes the precision and scale of the reconstructed scenes.

\section*{ACKNOWLEDGMENTS}

The authors would like to thank Tianji Jiang, Xi Xu, Jiadong Tang, Zhaoxiang Liang, Xihan Wang, Dianyi Yang, bohan Ren, Yixian Wang, and all other members of the ININ Lab of the Beijing Institute of Technology
for their contribution to this work.

\printbibliography

@inproceedings{ref1,
author={Mildenhall, Ben and Srinivasan, Pratul P. and Tancik, Matthew and Barron, Jonathan T. and Ramamoorthi, Ravi and Ng, Ren},
editor={Vedaldi, Andrea
and Bischof, Horst
and Brox, Thomas
and Frahm, Jan-Michael},
title={NeRF: Representing Scenes as Neural Radiance Fields for View Synthesis},
booktitle={Computer Vision -- ECCV 2020},
year={2020},
publisher={Springer International Publishing},
pages={405--421},
}

@article{ref2,
author = {Mildenhall, Ben and Srinivasan, Pratul P. and Ortiz-Cayon, Rodrigo and Kalantari, Nima Khademi and Ramamoorthi, Ravi and Ng, Ren and Kar, Abhishek},
title = {Local light field fusion: practical view synthesis with prescriptive sampling guidelines},
year = {2019},
issue_date = {August 2019},
publisher = {Association for Computing Machinery},
volume = {38},
number = {4},
journal = {ACM Trans. Graph.},
numpages = {14}
}

@inproceedings{ref3,
author = {Liu, Lingjie and Gu, Jiatao and Lin, Kyaw Zaw and Chua, Tat-Seng and Theobalt, Christian},
title = {Neural sparse voxel fields},
year = {2020},
publisher = {Curran Associates Inc},
booktitle = {Proceedings of the 34th International Conference on Neural Information Processing Systems},
numpages = {13},
location = {Vancouver, BC, Canada},
series = {NIPS'20}
}

@inproceedings{ref4,
  author={Barron, Jonathan T. and Mildenhall, Ben and Verbin, Dor and Srinivasan, Pratul P. and Hedman, Peter},
  booktitle={2022 IEEE/CVF Conference on Computer Vision and Pattern Recognition (CVPR)}, 
  title={Mip-NeRF 360: Unbounded Anti-Aliased Neural Radiance Fields}, 
  year={2022},
  volume={},
  number={},
  pages={5460-5469}
}

@inproceedings{ref5,
  author={Schönberger, Johannes L. and Frahm, Jan-Michael},
  booktitle={2016 IEEE Conference on Computer Vision and Pattern Recognition (CVPR)}, 
  title={Structure-from-Motion Revisited}, 
  year={2016},
  volume={},
  number={},
  pages={4104-4113}
}

@book{ref8,
  title={Three-dimensional computer vision: a geometric viewpoint},
  author={Faugeras, Olivier},
  year={1993},
  publisher={MIT press}
}

@article{ref9,
  author={Tsai, R.},
  journal={IEEE Journal on Robotics and Automation}, 
  title={A versatile camera calibration technique for high-accuracy 3D machine vision metrology using off-the-shelf TV cameras and lenses}, 
  year={1987},
  volume={3},
  number={4},
  pages={323-344}
}

@inproceedings{ref10,
  author={Zhengyou Zhang},
  booktitle={Proceedings of the Seventh IEEE International Conference on Computer Vision}, 
  title={Flexible camera calibration by viewing a plane from unknown orientations}, 
  year={1999},
  volume={1},
  number={},
  pages={666-673 vol.1}
}

@ARTICLE{ref11,
  author={Lee, Hyunjoon and Shechtman, Eli and Wang, Jue and Lee, Seungyong},
  journal={IEEE Transactions on Pattern Analysis and Machine Intelligence}, 
  title={Automatic Upright Adjustment of Photographs With Robust Camera Calibration}, 
  year={2014},
  volume={36},
  number={5},
  pages={833-844}
}

@inproceedings{ref12,
  author={Zhai, Menghua and Workman, Scott and Jacobs, Nathan},
  booktitle={2016 IEEE Conference on Computer Vision and Pattern Recognition (CVPR)}, 
  title={Detecting Vanishing Points Using Global Image Context in a Non-ManhattanWorld}, 
  year={2016},
  volume={},
  number={},
  pages={5657-5665}
}

@inproceedings{ref13,
author="Chen, Qian
and Wu, Haiyuan
and Wada, Toshikazu",
editor="Pajdla, Tom{\'a}{\v{s}}
and Matas, Ji{\v{r}}{\'i}",
title="Camera Calibration with Two Arbitrary Coplanar Circles",
booktitle="Computer Vision - ECCV 2004",
year="2004",
publisher="Springer Berlin Heidelberg",
pages="521--532"
}

@inproceedings{ref14,
author = {Bogdan, Oleksandr and Eckstein, Viktor and Rameau, Francois and Bazin, Jean-Charles},
title = {DeepCalib: a deep learning approach for automatic intrinsic calibration of wide field-of-view cameras},
year = {2018},
publisher = {Association for Computing Machinery},
address = {New York, NY, USA},
booktitle = {Proceedings of the 15th ACM SIGGRAPH European Conference on Visual Media Production},
numpages = {10}
}

@article{ref15,
  author={Hold-Geoffroy, Yannick and Piché-Meunier, Dominique and Sunkavalli, Kalyan and Bazin, Jean-Charles and Rameau, François and Lalonde, Jean-François},
  journal={IEEE Transactions on Pattern Analysis and Machine Intelligence}, 
  title={A Perceptual Measure for Deep Single Image Camera and Lens Calibration}, 
  year={2023},
  volume={45},
  number={9},
  pages={10603-10614},
}

@inbook{ref16,
  author={Landy, Michael and Movshon, J. Anthony},
  booktitle={Computational Models of Visual Processing}, 
  title={Shape from X: Psychophysics and Computation}, 
  year={1991},
  volume={},
  number={},
  pages={305-330},
}

@inproceedings {ref20,
author = {W. Ye and X. Lan and S. Chen and Y. Ming and X. Yu and H. Bao and Z. Cui and G. Zhang},
booktitle = {2023 IEEE/CVF Conference on Computer Vision and Pattern Recognition (CVPR)},
title = {PVO: Panoptic Visual Odometry},
year = {2023},
pages = {9579-9589},
publisher = {IEEE Computer Society}
}

@inproceedings{ref24,
author = {Yen-Chen, Lin and Florence, Pete and Barron, Jonathan T. and Rodriguez, Alberto and Isola, Phillip and Lin, Tsung-Yi},
title = {iNeRF: Inverting Neural Radiance Fields for Pose Estimation},
year = {2021},
publisher = {IEEE Press},
booktitle = {2021 IEEE/RSJ International Conference on Intelligent Robots and Systems (IROS)},
pages = {1323–1330},
numpages = {8}
}

@inproceedings{ref26,
author = {R. Martin-Brualla and N. Radwan and M. M. Sajjadi and J. T. Barron and A. Dosovitskiy and D. Duckworth},
booktitle = {2021 IEEE/CVF Conference on Computer Vision and Pattern Recognition (CVPR)},
title = {NeRF in the Wild: Neural Radiance Fields for Unconstrained Photo Collections},
year = {2021},
pages = {7206-7215},
publisher = {IEEE Computer Society}
}

@INPROCEEDINGS {ref31,
author = {J. T. Barron and B. Mildenhall and D. Verbin and P. P. Srinivasan and P. Hedman},
booktitle = {2023 IEEE/CVF International Conference on Computer Vision (ICCV)},
title = {Zip-NeRF: Anti-Aliased Grid-Based Neural Radiance Fields},
year = {2023},
pages = {19640-19648},
publisher = {IEEE Computer Society}
}

@article{ref32,
    author = {Thomas M\"uller and Alex Evans and Christoph Schied and Alexander Keller},
    title = {Instant Neural Graphics Primitives with a Multiresolution Hash Encoding},
    journal = {ACM Trans. Graph.},
    issue_date = {July 2022},
    volume = {41},
    number = {4},
    month = jul,
    year = {2022},
    pages = {102:1--102:15},
    articleno = {102},
    numpages = {15},
    publisher = {ACM},
    address = {New York, NY, USA}
}

@INPROCEEDINGS{ref34,
  author={Wu, Changchang},
  booktitle={2013 International Conference on 3D Vision - 3DV 2013}, 
  title={Towards Linear-Time Incremental Structure from Motion}, 
  year={2013},
  volume={},
  number={},
  pages={127-134},
}

@INPROCEEDINGS{ref35,
  author={Radenovic, Filip and Schönberger, Johannes L. and Ji, Dinghuang and Frahm, Jan-Michael and Chum, Ondrej and Matas, Jirí},
  booktitle={2016 IEEE Conference on Computer Vision and Pattern Recognition (CVPR)}, 
  title={From Dusk Till Dawn: Modeling in the Dark}, 
  year={2016},
  volume={},
  number={},
  pages={5488-5496}
}

@INPROCEEDINGS{ref36,
  author={Li, Zhengqi and Niklaus, Simon and Snavely, Noah and Wang, Oliver},
  booktitle={2021 IEEE/CVF Conference on Computer Vision and Pattern Recognition (CVPR)}, 
  title={Neural Scene Flow Fields for Space-Time View Synthesis of Dynamic Scenes}, 
  year={2021},
  volume={},
  number={},
  pages={6494-6504}
}

@InProceedings{ref37,
author="Chng, Shin-Fang
and Ramasinghe, Sameera
and Sherrah, Jamie
and Lucey, Simon",
editor="Avidan, Shai
and Brostow, Gabriel
and Ciss{\'e}, Moustapha
and Farinella, Giovanni Maria
and Hassner, Tal",
title="Gaussian Activated Neural Radiance Fields for High Fidelity Reconstruction and Pose Estimation",
booktitle="Computer Vision -- ECCV 2022",
year="2022",
publisher="Springer Nature Switzerland",
pages="264--280"
}

@INPROCEEDINGS{ref38,
  author={Lin, Chen-Hsuan and Ma, Wei-Chiu and Torralba, Antonio and Lucey, Simon},
  booktitle={2021 IEEE/CVF International Conference on Computer Vision (ICCV)}, 
  title={BARF: Bundle-Adjusting Neural Radiance Fields}, 
  year={2021},
  volume={},
  number={},
  pages={5721-5731}
}

@misc{ref39,
      title={NeRF--: Neural Radiance Fields Without Known Camera Parameters}, 
      author={Zirui Wang and Shangzhe Wu and Weidi Xie and Min Chen and Victor Adrian Prisacariu},
      year={2022},
      eprint={2102.07064},
      archivePrefix={arXiv},
      primaryClass={cs.CV}
}

@inproceedings{ref40,
  title={SiNeRF: Sinusoidal Neural Radiance Fields for Joint Pose Estimation and Scene Reconstruction},
  author={Yitong Xia and Hao Tang and Radu Timofte and Luc Van Gool},
  booktitle={British Machine Vision Conference},
  year={2022}
}

@article{ref41,
  title={Self-Calibrating Neural Radiance Fields},
  author={Yoonwoo Jeong and Seokjun Ahn and Christopher Bongsoo Choy and Anima Anandkumar and Minsu Cho and Jaesik Park},
  journal={2021 IEEE/CVF International Conference on Computer Vision (ICCV)},
  year={2021},
  pages={5826-5834}
}

@INPROCEEDINGS{ref42,
  author={Park, Keunhong and Sinha, Utkarsh and Barron, Jonathan T. and Bouaziz, Sofien and Goldman, Dan B and Seitz, Steven M. and Martin-Brualla, Ricardo},
  booktitle={2021 IEEE/CVF International Conference on Computer Vision (ICCV)}, 
  title={Nerfies: Deformable Neural Radiance Fields}, 
  year={2021},
  volume={},
  number={},
  pages={5845-5854}
}

@inproceedings{ref43,
  title={SAPE: Spatially-Adaptive Progressive Encoding for Neural Optimization},
  author={Amir Hertz and Or Perel and Raja Giryes and Olga Sorkine-Hornung and Daniel Cohen-Or},
  booktitle={Neural Information Processing Systems},
  year={2021}
}

@inproceedings{ref44,
 author = {Vaswani, Ashish and Shazeer, Noam and Parmar, Niki and Uszkoreit, Jakob and Jones, Llion and Gomez, Aidan N and Kaiser, \L ukasz and Polosukhin, Illia},
 booktitle = {Advances in Neural Information Processing Systems},
 editor = {I. Guyon and U. Von Luxburg and S. Bengio and H. Wallach and R. Fergus and S. Vishwanathan and R. Garnett},
 pages = {},
 publisher = {Curran Associates, Inc.},
 title = {Attention is All you Need},
 volume = {30},
 year = {2017}
}

@INPROCEEDINGS {ref45,
author = {W. Bian and Z. Wang and K. Li and J. Bian},
booktitle = {2023 IEEE/CVF Conference on Computer Vision and Pattern Recognition (CVPR)},
title = {NoPe-NeRF: Optimising Neural Radiance Field with No Pose Prior},
year = {2023},
pages = {4160-4169},
publisher = {IEEE Computer Society}
}

@INPROCEEDINGS {ref46,
author = {P. Truong and M. Rakotosaona and F. Manhardt and F. Tombari},
booktitle = {2023 IEEE/CVF Conference on Computer Vision and Pattern Recognition (CVPR)},
title = {SPARF: Neural Radiance Fields from Sparse and Noisy Poses},
year = {2023},
pages = {4190-4200},
publisher = {IEEE Computer Society}
}

@INPROCEEDINGS {ref47,
author = {Y. Chen and G. Lee},
booktitle = {2023 IEEE/CVF Conference on Computer Vision and Pattern Recognition (CVPR)},
title = {DBARF: Deep Bundle-Adjusting Generalizable Neural Radiance Fields},
year = {2023},
volume = {},
issn = {},
pages = {24-34},
publisher = {IEEE Computer Society}
}

@INPROCEEDINGS {ref48,
author = {Q. Wang and Z. Wang and K. Genova and P. Srinivasan and H. Zhou and J. T. Barron and R. Martin-Brualla and N. Snavely and T. Funkhouser},
booktitle = {2021 IEEE/CVF Conference on Computer Vision and Pattern Recognition (CVPR)},
title = {IBRNet: Learning Multi-View Image-Based Rendering},
year = {2021},
volume = {},
issn = {},
pages = {4688-4697},
publisher = {IEEE Computer Society}
}

@INPROCEEDINGS{ref49,
  author={Yu, Alex and Ye, Vickie and Tancik, Matthew and Kanazawa, Angjoo},
  booktitle={2021 IEEE/CVF Conference on Computer Vision and Pattern Recognition (CVPR)}, 
  title={pixelNeRF: Neural Radiance Fields from One or Few Images}, 
  year={2021},
  volume={},
  number={},
  pages={4576-4585}
}

@INPROCEEDINGS{ref50,
  author={Meng, Quan and Chen, Anpei and Luo, Haimin and Wu, Minye and Su, Hao and Xu, Lan and He, Xuming and Yu, Jingyi},
  booktitle={2021 IEEE/CVF International Conference on Computer Vision (ICCV)}, 
  title={GNeRF: GAN-based Neural Radiance Field without Posed Camera}, 
  year={2021},
  volume={},
  number={},
  pages={6331-6341}
}

@INPROCEEDINGS{ref51,
  author={Deng, Jia and Dong, Wei and Socher, Richard and Li, Li-Jia and Kai Li and Li Fei-Fei},
  booktitle={2009 IEEE Conference on Computer Vision and Pattern Recognition}, 
  title={ImageNet: A large-scale hierarchical image database}, 
  year={2009},
  volume={},
  number={},
  pages={248-255},
  doi={10.1109/CVPR.2009.5206848}
}

@BOOK{ref52,
  title={Multiple View Geometry in Computer Vision},
  author={Hartley, Richard I and Zisserman, Andrew},
  year={2003},
  publisher={Cambridge University Press},
  isbn={0521540518},
  edition={2}
}

@INPROCEEDINGS{ref54,
  title={Local-to-global registration for bundle-adjusting neural radiance fields},
  author={Chen, Yue and Chen, Xingyu and Wang, Xuan and Zhang, Qi and Guo, Yu and Shan, Ying and Wang, Fei},
  booktitle={Proceedings of the IEEE/CVF Conference on Computer Vision and Pattern Recognition},
  pages={8264--8273},
  year={2023}
}

@article{ref55,
author = {Park, Keunhong and Henzler, Philipp and Mildenhall, Ben and Barron, Jonathan T. and Martin-Brualla, Ricardo},
title = {CamP: Camera Preconditioning for Neural Radiance Fields},
year = {2023},
issue_date = {December 2023},
publisher = {Association for Computing Machinery},
volume = {42},
number = {6},
journal = {ACM Trans. Graph.},
numpages = {11},
}

@article{ref56,
  title={3D Gaussian Splatting for Real-Time Radiance Field Rendering.},
  author={Kerbl, Bernhard and Kopanas, Georgios and Leimk{\"u}hler, Thomas and Drettakis, George},
  journal={ACM Trans. Graph.},
  volume={42},
  number={4},
  pages={139--1},
  year={2023}
}

@article{ref57,
  title={Instantsplat: Unbounded sparse-view pose-free gaussian splatting in 40 seconds},
  author={Fan, Zhiwen and Cong, Wenyan and Wen, Kairun and Wang, Kevin and Zhang, Jian and Ding, Xinghao and Xu, Danfei and Ivanovic, Boris and Pavone, Marco and Pavlakos, Georgios and others},
  journal={arXiv preprint arXiv:2403.20309},
  year={2024}
}

@INPROCEEDINGS{ref58,
  author={Fu, Yang and Wang, Xiaolong and Liu, Sifei and Kulkarni, Amey and Kautz, Jan and Efros, Alexei A.},
  booktitle={2024 IEEE/CVF Conference on Computer Vision and Pattern Recognition (CVPR)}, 
  title={COLMAP-Free 3D Gaussian Splatting}, 
  year={2024},
  volume={},
  number={},
  pages={20796-20805},
  doi={10.1109/CVPR52733.2024.01965}
}

@INPROCEEDINGS{ref59,
  author={Yan, Chi and Qu, Delin and Xu, Dan and Zhao, Bin and Wang, Zhigang and Wang, Dong and Li, Xuelong},
  booktitle={2024 IEEE/CVF Conference on Computer Vision and Pattern Recognition (CVPR)}, 
  title={GS-SLAM: Dense Visual SLAM with 3D Gaussian Splatting}, 
  year={2024},
  volume={},
  number={},
  pages={19595-19604},
  doi={10.1109/CVPR52733.2024.01853}
}

@INPROCEEDINGS{ref60,
  author={Keetha, Nikhil and Karhade, Jay and Jatavallabhula, Krishna Murthy and Yang, Gengshan and Scherer, Sebastian and Ramanan, Deva and Luiten, Jonathon},
  booktitle={2024 IEEE/CVF Conference on Computer Vision and Pattern Recognition (CVPR)}, 
  title={SplaTAM: Splat, Track $\&$ Map 3D Gaussians for Dense RGB-D SLAM}, 
  year={2024},
  volume={},
  number={},
  pages={21357-21366},
  doi={10.1109/CVPR52733.2024.02018}
}

@INPROCEEDINGS{ref61,
  author={Wang, Shuzhe and Leroy, Vincent and Cabon, Yohann and Chidlovskii, Boris and Revaud, Jerome},
  booktitle={2024 IEEE/CVF Conference on Computer Vision and Pattern Recognition (CVPR)}, 
  title={DUSt3R: Geometric 3D Vision Made Easy}, 
  year={2024},
  volume={},
  number={},
  pages={20697-20709},
  doi={10.1109/CVPR52733.2024.01956}
}

@article{ref63,
  title={Humanrf: High-fidelity neural radiance fields for humans in motion},
  author={I{\c{s}}{\i}k, Mustafa and R{\"u}nz, Martin and Georgopoulos, Markos and Khakhulin, Taras and Starck, Jonathan and Agapito, Lourdes and Nie{\ss}ner, Matthias},
  journal={ACM Transactions on Graphics (TOG)},
  volume={42},
  number={4},
  pages={1--12},
  year={2023},
  publisher={ACM New York, NY, USA}
}

@inproceedings{ref64,
  title={Dna-rendering: A diverse neural actor repository for high-fidelity human-centric rendering},
  author={Cheng, Wei and Chen, Ruixiang and Fan, Siming and Yin, Wanqi and Chen, Keyu and Cai, Zhongang and Wang, Jingbo and Gao, Yang and Yu, Zhengming and Lin, Zhengyu and others},
  booktitle={Proceedings of the IEEE/CVF International Conference on Computer Vision},
  pages={19982--19993},
  year={2023}
}
\newpage
\section*{Supplementary Materials}
\subsection*{Supplementary performance in Real-World Datasets}

\begin{figure*}[htbp]
\centering
\includegraphics[width=\linewidth]{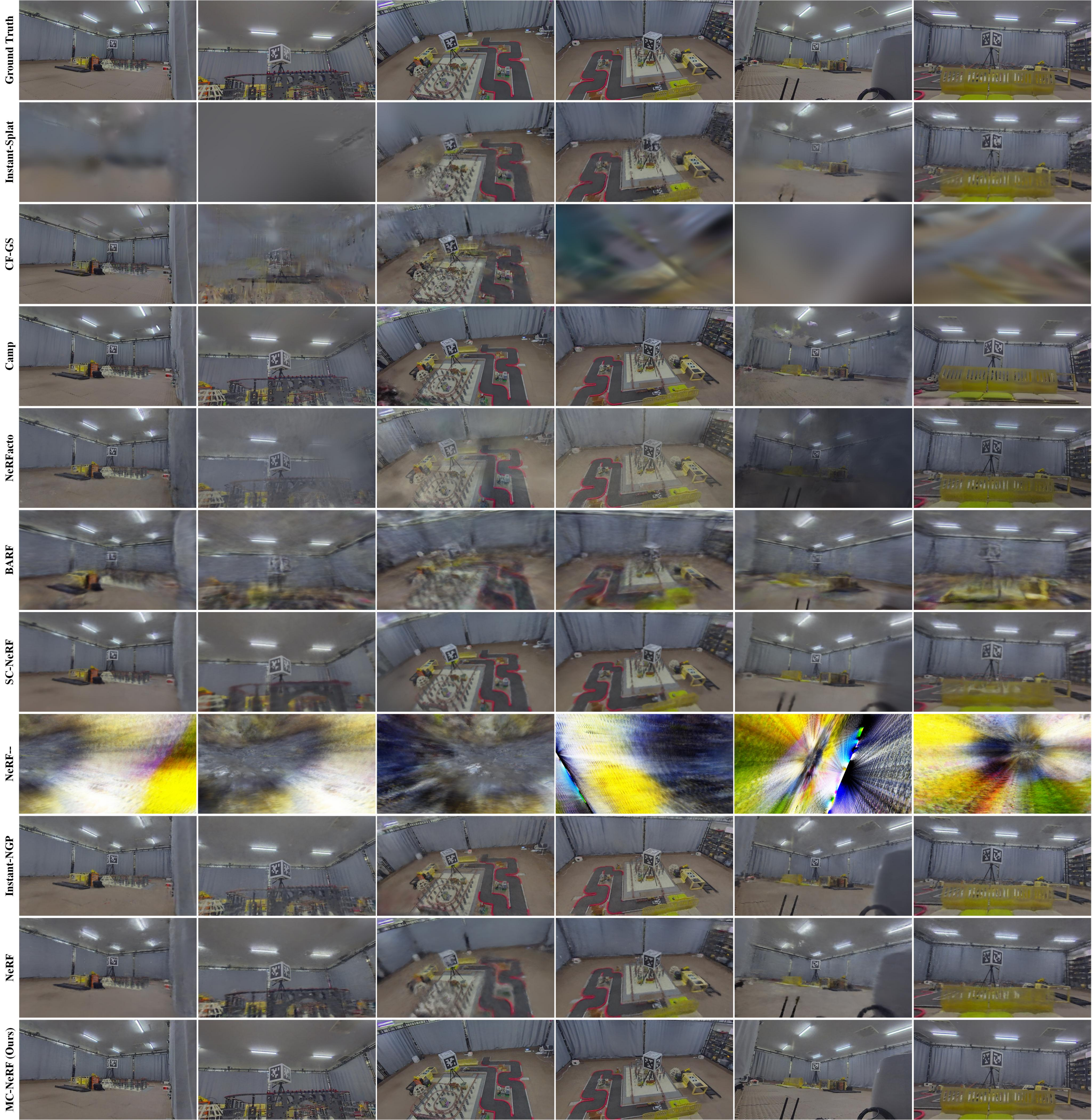}
\caption{Rendering results of different methods in real-world dataset $Lab2$. In our comparative experiments, InstantSplat and CF Gaussian autonomously estimate camera parameters without any provided initializations. Methods such as CAMP, NeRFacto, and BARF, which are initialized using camera parameters provided by Colmap, are limited to optimizing only the extrinsic parameters. In contrast, SC-NeRF and NeRF$--$, also initialized with Colmap, are capable of optimizing both intrinsic and extrinsic parameters. Instant-NGP, which lacks camera parameter optimization capabilities, serves as a baseline reference. MC-NeRF utilizes auxiliary images for initialization, and it is also capable of optimizing both intrinsic and extrinsic parameters. The results demonstrate that our method achieved the competitive rendering performance among all compared methods.}
\label{Fig.13}
\end{figure*}

\begin{figure*}[htbp]
\centering
\includegraphics[width=\linewidth]{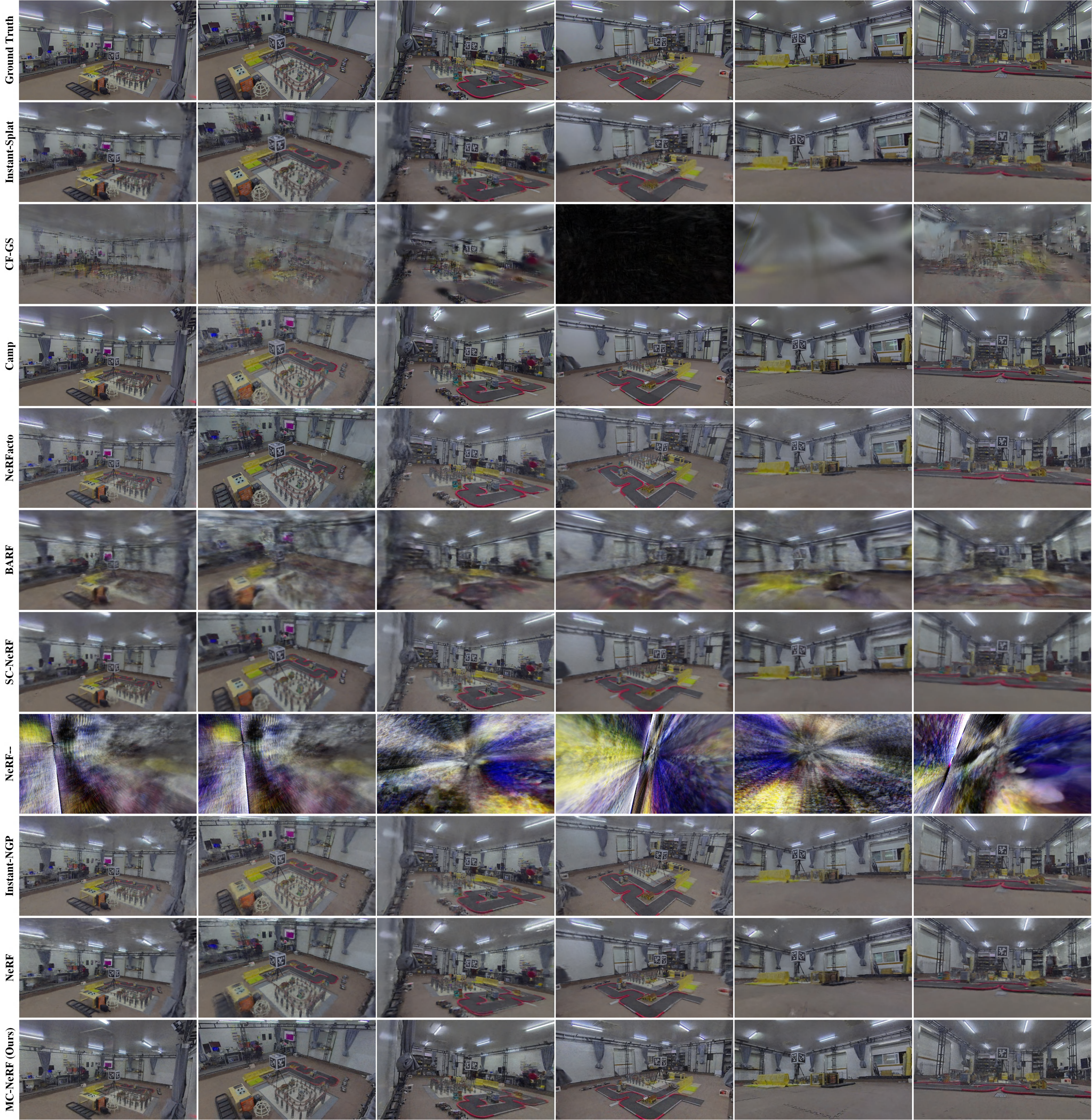}
\caption{Rendering results of different methods in real-world dataset $Lab3$. In our comparative experiments, InstantSplat and CF Gaussian autonomously estimate camera parameters without any provided initializations. Methods such as CAMP, NeRFacto, and BARF, which are initialized using camera parameters provided by Colmap, are limited to optimizing only the extrinsic parameters. In contrast, SC-NeRF and NeRF$--$, also initialized with Colmap, are capable of optimizing both intrinsic and extrinsic parameters. Instant-NGP, which lacks camera parameter optimization capabilities, serves as a baseline reference. MC-NeRF utilizes auxiliary images for initialization, and it is also capable of optimizing both intrinsic and extrinsic parameters. The results demonstrate that our method achieved the competitive rendering performance among all compared methods.}
\label{Fig.14}
\end{figure*}

In the original paper, EXP.4 in Sec 4.4 provided the rendering results for $Lab1$ Scene, while the rendering results for the other two scenes are shown in Fig. \ref{Fig.13} and Fig. \ref{Fig.14}. The comparative methods and metrics setup are consistent with Fig. 12 in the original paper. From the rendering results of the $Lab2$ and $Lab3$ scenes, it can be seen that our method achieves competitive performance across different scenes, further demonstrating its effectiveness."

\subsection*{Supplementary Experiment A: Fix Step VS Global optimization}

In this experiment, we aim to validate the effectiveness of the global optimization framework and compare its performance against the Fixed Step NeRF rendering (Intrinsic and extrinsic parameters estimation + NeRF, without other optimization). Fixed Step rendering simulates the typical steps involved in NeRF-based reconstruction using a multi-camera acquisition systems: Initially, camera intrinsic and extrinsic parameters are determined using calibration information. Subsequently, NeRF is trained using these fixed parameters to represent the 3D scene. Throughout the NeRF training phase, the camera parameters remain constant. 

\begin{figure*}[htbp]
\centering
\includegraphics[width=\linewidth]{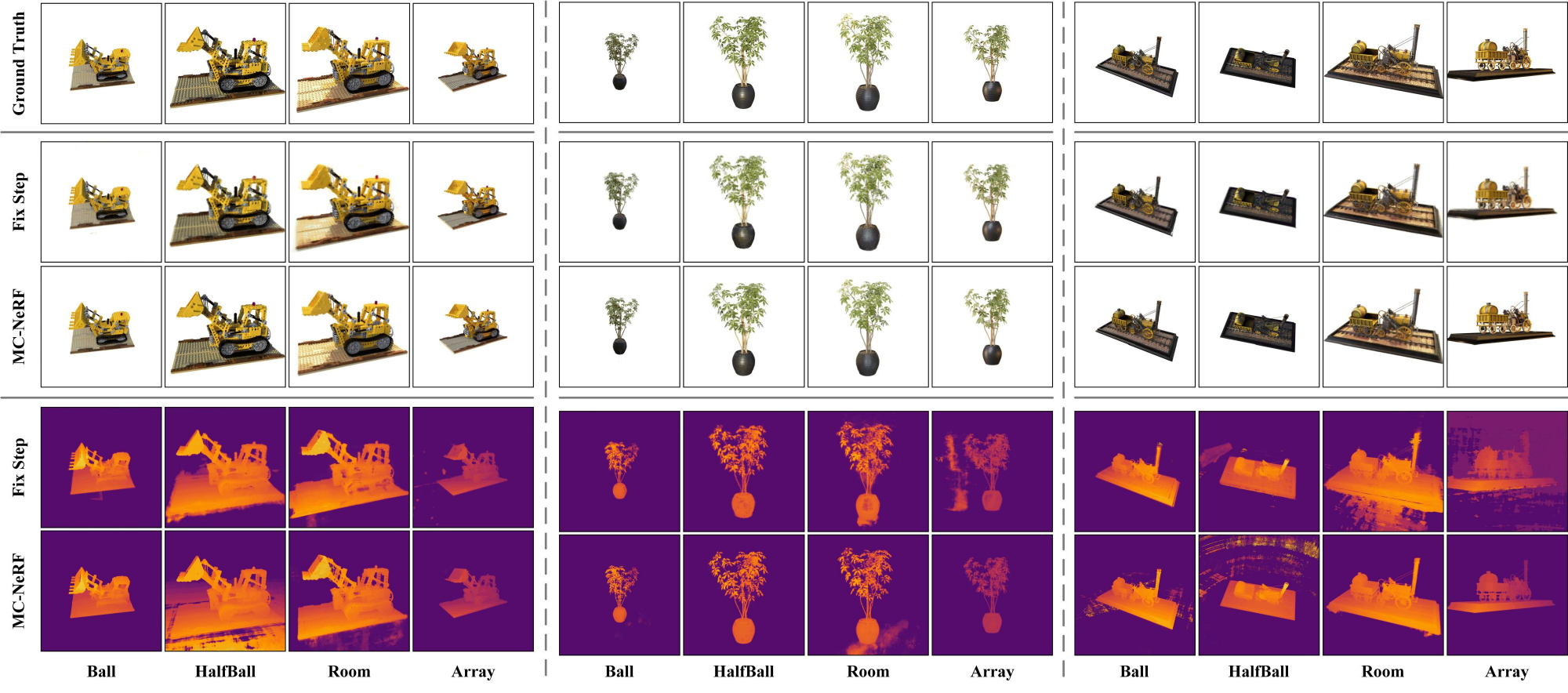}
\caption{Rendering results of Fix-Step NeRF and MC-NeRF. We visualize the ground truth images (top row), the rendered RGB images (middle row), and the predicted depth images (bottom row). Thanks to the global optimization, MC-NeRF exhibits improved rendering details and sharper object boundaries.}
\label{Fig.15}
\end{figure*}

The experimental results are presented in Fig.\hyperref[Fig.3]{\ref{Fig.15}}, Table \ref{Table 6}, and Table \ref{Table 7}. Regarding image rendering quality, in Fig.\hyperref[Fig.3]{\ref{Fig.15}}, we randomly selected three scenes rendered in four different styles. The global optimization method stands out by producing sharper object boundaries and capturing finer object details. Table \ref{Table 6} shows the rendering performance for 4 styles, and Table \ref{Table 7} shows the camera parameters estimation results. Firstly, MC-NeRF shows a significant performance improvement in Array Style, where it achieves the best scores across all three evaluation metrics. Secondly, we observed that MC-NeRF consistently outperforms in all LPIPS metrics. However, when it comes to PSNR and SSIM, except for the Array scene, MC-NeRF often has lower scores. We explored the reasons, and the detailed analyses and experiments are discussed in SUPPLEMENTARY EXPERIMENT D.

\subsection*{Supplementary Experiment B: Stacking Techniques}
\begin{figure*}[htbp]
\centering
\includegraphics[width=\linewidth]{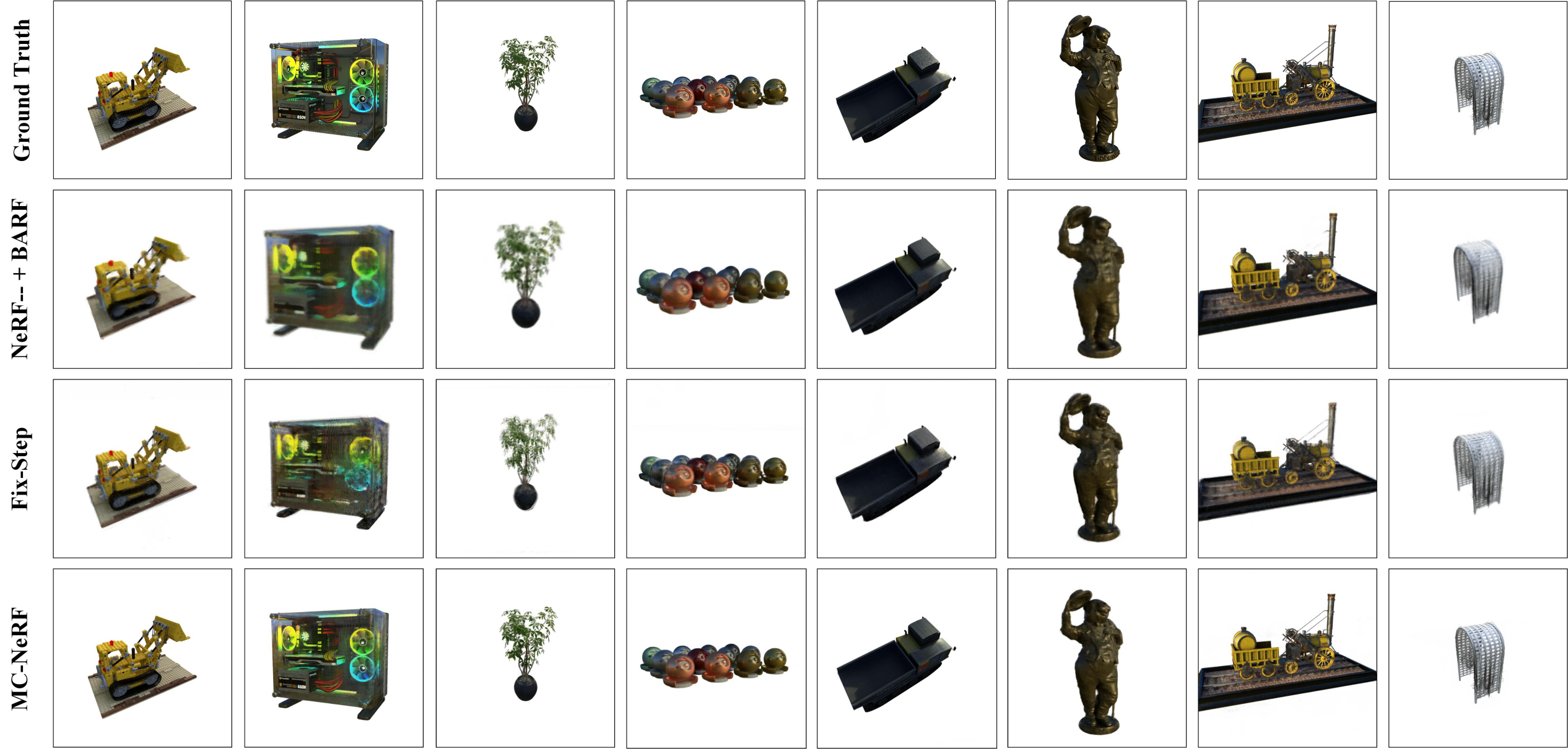}
\caption{Rendering results of BARF+NeRF$--$, Fix-Step NeRF and MC-NeRF. The first row represents the ground truth images, while the remaining rows display rendering results. The rendering results indicate that simply stacking methods without designing an optimized sequence does not lead to improved performance.}
\label{Fig.16}
\end{figure*}

In MC-NeRF, the NeRF component does not undergo significant modifications, the method for estimating camera parameters is derived from NeRF$--$ and BARF. However, simply stacking methods cannot achieve the effect of MC-NeRF, mainly for two reasons: First, the parameter coupling issue, which has already been emphasized in the original manuscript. For this part, we provide additional AprilTag information to ensure the parameters can be decoupled. Second, the training process needs to follow the training sequence diagram proposed in the original manuscript (Fig. 6: Training Sequence Diagram). Here, we conducted comparative experiments by simply stacking the techniques of BARF and NeRF$--$ and using the same camera parameter initialization process as MC-NeRF. The results are shown in Fig.\ref{Fig.16} and Table \ref{Table 8}.

As shown in Table \ref{Table 8}, Mix-Step represents the hybrid method of \textit{AprilTag + BARF + NeRF$--$}, and the comparative results demonstrate that simply stacking techniques without adhering to the aforementioned conditions cannot achieve the desired results.

\subsection*{Supplementary Experiment C: NeRF with Off-the-Shelf Libraries}

In addition to the comparison experiments with real-world datasets described in the original manuscript, we carried out further supplementary experiments to substantiate the performance of MC-NeRF in synthetic datasets. In these experiments, we employed camera parameters obtained using SFM (Structure from Motion) techniques from well-established libraries including COLMAP and Meshroom. These parameters are then used to train with NeRF. The count of usable images sourced from these libraries is presented in Table \ref{Table 9}.

Table \ref{Table 9} demonstrates that with synthetic datasets, SFM-based methods exhibit unreliable performance, leading to a significant loss of image data. This can be attributed to the simplistic background textures prevalent in synthetic data, coupled with the inherently limited number of images available. Conversely, the estimation method using AprilTag does not encounter these issues, which is one of the reasons we opted for this method.

\begin{figure*}[htbp]
\centering
\includegraphics[width=\linewidth]{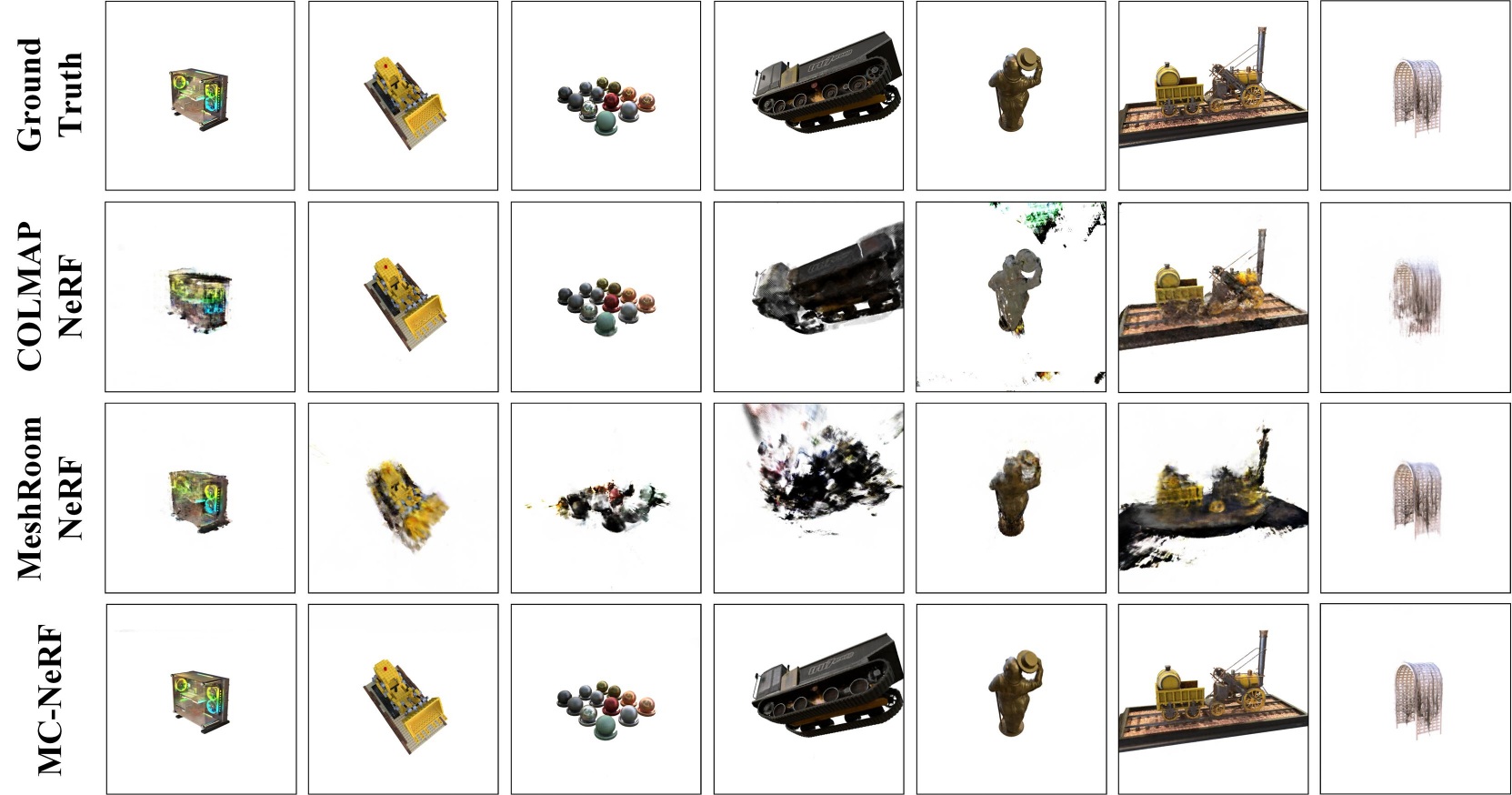}
\caption{Rendering results of COLMAP+NeRF, Meshroom+NeRF and MC-NeRF. The first row represents the ground truth images, while the remaining rows display rendering results. Based on the rendering results, SFM-based methods are not applicable across all scenes and use cases. In scenarios with sparse backgrounds or textures, the camera parameters obtained are often inaccurate or even fail entirely. In contrast, our method demonstrates greater generalization and reliability, though it does require additional image data, which increases the data collection workload.}
\label{Fig.17}
\end{figure*}

To ensure that the experimental results are meaningful, we restricted our training to scenes with more than 30 usable images. The rendering results illustrated in Fig. \ref{Fig.17} and Table \ref{Table 10}. From Fig. \ref{Fig.17}, it is seen that MC-NeRF achieves better rendering details, particularly in terms of object edge delineation. Furthermore, in quantitative results, MC-NeRF also outperforms the other two methods across all three metrics.

\subsection*{Supplementary Experiment D: PSNR and SSIM evaluation metrics analysis}

\begin{figure*}[htbp]
\centering
\includegraphics[width=\linewidth]{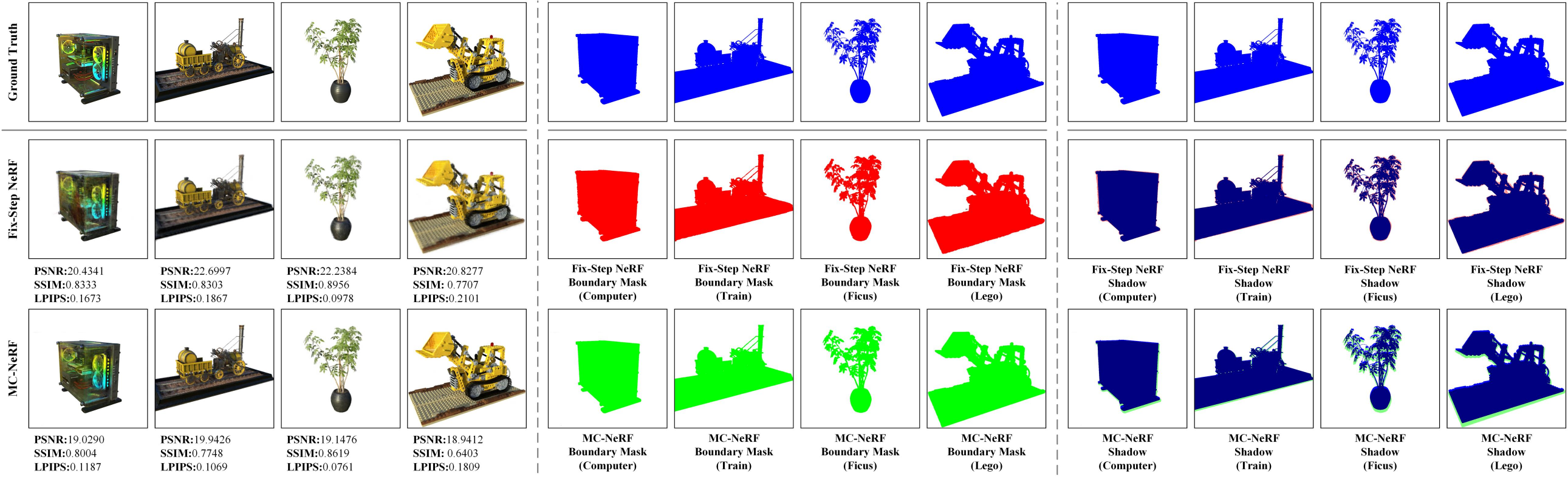}
\caption{Boundary alignment results for rendering objects. On the left, the rendering results and evaluation metrics for Fix-Step NeRF and MC-NeRF are displayed. In the middle section, the object boundary masks are presented. On the right, the comparison results between the object boundaries and the ground truth are shown. It can be observed that Fix-Step NeRF exhibits better object alignment performance compared to MC-NeRF. The more pixels contained within the mask, the more favorable it is for evaluation methods like PSNR, which involve pixel-to-pixel calculations.}
\label{Fig.18}
\end{figure*}

Although MC-NeRF achieved better rendering details, it scored lower on both PSNR and SSIM evaluation metrics. We believe this is related to the alignment of object boundaries, and the reason behind this phenomenon is the absence of enforced ray alignment constraints for rendering in MC-NeRF. Specifically, the camera parameters of Fix-Step NeRF are constant and remain stable during the rendering stage, which maintains a fixed ray distribution, and all fitting computations contained within NeRF's MLPs. It can be observed that to compensate for the ray distribution error resulting from inaccuracies in camera parameters, the rendered objects exhibit a multi-layer shadow structure. In Fig.\hyperref[Fig.18]{\ref{Fig.18}}, the boundaries of the computer show noticeable shadows, while those of MC-NeRF appear much clearer and sharper. This multi-layer shadow structure has more advantages for pixel-to-pixel calculation methods like PSNR. If the clear-edged object generated by MC-NeRF is not aligned with the ground truth, the PSNR scores can be easily affected by the background, resulting in lower scores. Although MC-NeRF has some drawbacks in terms of boundary alignment, we prioritize high-quality rendering. Additionally, based on the rendering images, we believe that this alignment error is tolerable and unlikely to be readily noticeable.

Moreover, in this experiment, we noted that achieving higher-quality rendering results does not necessarily correlate with higher PSNR and SSIM scores. Because even when the estimation results for both intrinsic and extrinsic parameters are highly accurate, their values are often challenging to align perfectly with ground truth. This leads to slight shifts of objects in the rendered image, which have a notable impact when calculating pixel-wise differences for PSNR and SSIM. However, the LPIPS can still objectively reflect rendering performance even in such cases.

\subsection{*Supplementary Experiment E: Unique-Camera VS Multi-Cameras}

In this experiment, we aim to explore whether NeRF can simultaneously utilize multiple intrinsic parameters, a foundational prerequisite for our method. We select cameras with FOVs of 40, 50, 60, 70, and 80 degrees to generate a mixed dataset. NeRF is then trained on this dataset with the provided ground-truth intrinsic and extrinsic parameters. Simultaneously, we conduct training of NeRF with each FOV data independently to compare the rendering performance against NeRF trained on the mixed dataset. Hence, this experiment comprises a total of 6 datasets: one mixed dataset and five independent datasets, each containing 84 images with identical extrinsic parameters across all datasets.

\begin{figure*}[htbp]
\centering
\includegraphics[width=\linewidth]{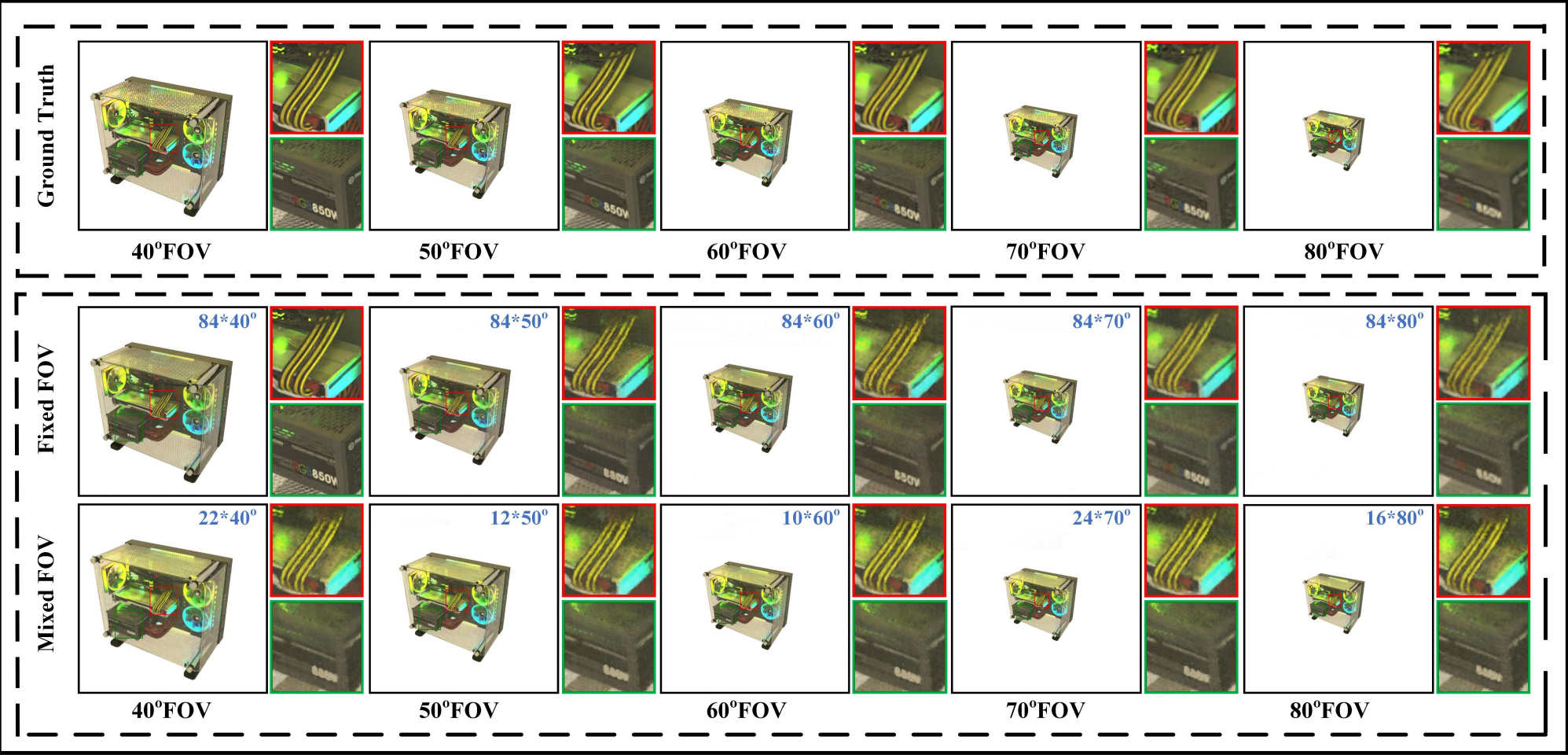}
\caption{Comparison of rendering performance in NeRF with mixed intrinsic dataset and single intrinsic dataset. The top row shows ground-truth images captured at varying FOVs, indicated below each image. The second row displays results produced by the NeRF model trained on each fixed FOV dataset, and the third row exhibits results from the NeRF model trained on a mixed dataset. The numbers in the top-right corner of each result represent the number of images in the dataset with the current FOV. In the mixed dataset, there are 22, 12, 10, 24, and 16 images for FOVs of 40, 50, 60, 70, and 80 degrees, respectively.}
\label{Fig.19}
\end{figure*}

The comparison results can be observed in Fig.\hyperref[Fig.19]{\ref{Fig.19}} and Table \ref{Table 11}. The $Distance$ item in Table \ref{Table 11} indicates the absolute difference between their respective indicators. As evident from both Table \ref{Table 11} and Fig.\hyperref[Fig.19]{\ref{Fig.19}}, the models trained with different datasets exhibit nearly equivalent performance. This observation has shown the capability of NeRF that can effectively process multi-camera images, when accurate camera intrinsic and extrinsic parameters are provided.

In summary, with accurate camera parameters provided, NeRF can work efficiently when each image corresponds to different intrinsic parameters. We believe that the challenge posed by multiple intrinsic parameters to NeRF is how to obtain accurate intrinsic parameters for each individual camera. The function of intrinsic and extrinsic parameters is to establish a mapping between image pixels and spatial rays. The accuracy and correctness of this mapping serve as vital prerequisites for NeRF, irrespective of whether the camera intrinsics are the same or equal.

\begin{table*}[htbp]
    \caption{Quantitative rendering results of Fix-Step NeRF and MC-NeRF on all scenes. From the table, it is evident that while MC-NeRF does not perform well in terms of PSNR and SSIM, it significantly outperforms the Fix-Step method in LPIPS. We analyzed the reasons behind this unexpected result, which are discussed in detail in SUPPLEMENTARY EXPERIMENT D.}
    \centering
    \renewcommand\arraystretch{1.4}
        \resizebox{\textwidth}{!}{
    \begin{tabular}{c|c|cccc|cccc|cccc}
    \toprule
    & \multirow{2}{*}{Scene} & \multicolumn{4}{c|}{PSNR$\uparrow$} & \multicolumn{4}{c|}{SSIM$\uparrow$} & \multicolumn{4}{c}{LPIPS$\downarrow$}\\
    & & Ball & HalfBall & Room & Array & Ball & HalfBall & Room & Array & Ball & HalfBall & Room & Array \\
    \hline
    \multirow{9}{*}{Fix-Step} & Computer & 25.1685 & 25.3800 & 20.6838 & 16.6674 & 0.9230 & 0.9088 & 0.8869 & 0.8838 & 0.0882 & 0.1020 & 0.1419 & 0.1571 \\
    & Ficus & 26.9503 & 25.9101 & 25.3156 & 17.2587 & 0.9539 & 0.9427 & 0.9313 & 0.8893 & 0.0431 & 0.0542 & 0.0762 & 0.1590 \\
    & Gate & 28.9521 & 28.9134 & 29.8277 & 18.1242 & 0.9416 & 0.9348 & 0.9341 & 0.8853 & 0.0542 & 0.0633 & 0.0782 & 0.1515 \\
    & Lego & 25.5330 & 25.1424 & 24.9465 & 15.6066 & 0.9219 & 0.8822 & 0.8742 & 0.8672 & 0.0998 & 0.1106 & 0.1214 & 0.1897 \\
    & Materials & 26.7998 & 25.8398 & 25.8738 & 16.3879 & 0.9443 & 0.9504 & 0.9462 & 0.8953 & 0.0666 & 0.0543 & 0.0634 & 0.1618 \\
    & Snowtruck & 25.1714 & 25.1153 & 23.7291 & 14.7770 & 0.9166 & 0.9114 & 0.8989 & 0.8770 & 0.0959 & 0.1093 & 0.1287 & 0.1728 \\
    & Statue & 30.5160 & 31.5379 & 24.4189 & 16.0766 & 0.9710 & 0.9694 & 0.9380 & 0.8938 & 0.0383 & 0.0420 & 0.0954 & 0.1571 \\
    & Train & 24.3452 & 24.1038 & 22.1991 & 15.0438 & 0.9240 & 0.8779 & 0.8578 & 0.8768 & 0.0876 & 0.1301 & 0.1466 & 0.1807 \\
    \cline{2-14}
    & Mean & 26.6795 & 26.4928 & 24.6243 & 16.2428 & 0.9370 & 0.9222 & 0.9084 & 0.8836 & 0.0717 & 0.0832 & 0.1065 & 0.1662 \\
    \midrule
    \multirow{9}{*}{MC-NeRF} & Computer &  24.2311& 23.0962 & 23.6603 &24.6946 &  0.9217 &  0.8876 & 0.9008 & 0.9395 & 0.0794 & 0.0933 & 0.1004 & 0.0444 \\
    & Ficus &  24.9479 &  24.7662 &  26.5778 & 24.2929 & 0.9488 &  0.9070 &  0.9386 & 0.9504 & 0.0336 &  0.0519 &  0.0449 & 0.0387 \\
    & Gate & 27.9102  & 26.0876  & 29.3081 & 26.7734 & 0.9333  & 0.9131 & 0.9250 & 0.9356 & 0.0519  & 0.0525 & 0.0602 & 0.0439\\
    & Lego & 24.1226 & 23.4818 & 23.3858 & 24.3502 & 0.9227 & 0.8682 & 0.8837 & 0.9629 & 0.0620 & 0.0901 & 0.0874 & 0.0343 \\
    & Materials & 26.8742 & 26.6170 & 27.1323 & 25.0106 &  0.9484 & 0.9488 & 0.9529 & 0.9601 & 0.0321 & 0.0423 & 0.0294& 0.0397 \\
    & Snowtruck & 24.8742 & 24.0678 & 23.4415 & 23.3363 & 0.9211 & 0.8882 & 0.8937 & 0.9375 & 0.0433 & 0.0882 & 0.0874 & 0.0451 \\
    & Statue & 28.6882 & 29.2767 & 28.7953 & 24.6619 & 0.9731 & 0.9670 & 0.9612 & 0.9651 & 0.0303 & 0.0301 & 0.0398 & 0.0281\\
    & Train & 23.3830 & 23.5271 & 22.2576 & 23.4073 & 0.9311 & 0.8773 & 0.8589 & 0.9395 & 0.0454 & 0.0991 & 0.0835 & 0.0507\\
    \cline{2-14}
    & Mean & 25.6289 & 25.1151 & 25.5698 & 24.5659 & 0.9375 & 0.9072 & 0.9144 & 0.9488 & 0.0473 & 0.0684 & 0.0666 & 0.0406 \\
    \bottomrule
    \end{tabular}}
    \label{Table 6}
\end{table*}  

\begin{table*}[htbp]
    \caption{Camera parameters estimation results of Fix-Step NeRF and MC-NeRF on all scenes. $Loss\_K$, $Loss\_R$, and $Loss\_T$ represent the average loss of all cameras in the rendering scene for the intrinsic parameters, rotation vectors, and translation vectors, respectively.}
    \centering
    \renewcommand\arraystretch{1.4}
        \resizebox{\textwidth}{!}{
    \begin{tabular}{c|c|cccc|cccc|cccc}
    \toprule
    & \multirow{2}{*}{Scene} & \multicolumn{4}{c|}{$Loss\_K$$\downarrow$} & \multicolumn{4}{c|}{$Loss\_R$$\downarrow$} & \multicolumn{4}{c}{$Loss\_T$$\downarrow$}\\
    & & Ball & HalfBall & Room & Array & Ball & HalfBall & Room & Array & Ball & HalfBall & Room & Array \\
    \hline
    \multirow{9}{*}{Fix-Step} & Computer & 12.0723 & 11.2606 & 17.6997 & 28.6428 & 0.0131 & 0.0124 & 0.0309 & 0.0157 & 0.0806 & 0.0830 & 0.1530 & 0.2616 \\
    & Ficus & 10.2547 & 11.0173 & 16.9588 & 28.9794 & 0.0147 & 0.0128 & 0.0276 & 0.0148 & 0.0802 & 0.0925 & 0.1507 & 0.2714 \\
    & Gate & 10.0650 & 11.0287 & 17.4918 & 26.3233 & 0.0134 & 0.0138 & 0.0344 & 0.0157 & 0.0781 & 0.0797 & 0.1604 & 0.2433 \\
    & Lego & 10.3323 & 11.0146 & 17.0964 & 24.9460 & 0.0135 & 0.0163 & 0.0315 & 0.0156 & 0.0787 & 0.0830 & 0.1577 & 0.2505 \\
    & Materials & 11.3069 & 10.8165 & 15.3205 & 27.4074 & 0.0157 & 0.0148 & 0.0300 & 0.0193 & 0.1005 & 0.0860 & 0.1458 & 0.2704 \\
    & Snowtruck & 11.7263 & 10.9522 & 13.9390 & 29.3079 & 0.0131 & 0.0136 & 0.0270 & 0.0141 & 0.0811 & 0.0727 & 0.1106 & 0.2619 \\
    & Statue & 10.3607 & 10.0933 & 17.1346 & 27.5365 & 0.0123 & 0.0104 & 0.0352 & 0.0149 & 0.0775 & 0.0774 & 0.1480 & 0.2569 \\
    & Train & 11.5821 & 10.2598 & 17.1833 & 27.8020 & 0.0136 & 0.0139 & 0.0334 & 0.0164 & 0.0862 & 0.0762 & 0.1630 & 0.2595 \\
    \cline{2-14}
    & Mean & 10.9625 & 10.8054 & 16.6030 & 27.6182 & 0.0137 & 0.0135 & 0.0313 & 0.0158 & 0.0829 & 0.0813 & 0.1487 & 0.2594 \\
    \midrule
    \multirow{9}{*}{MC-NeRF} & Computer &1.2758& 5.4864  &6.3094& 5.0589 & 0.0033 & 0.0101 &0.0090 & 0.0057 & 0.0231 &0.0398  &0.0696& 0.0549 \\
    & Ficus & 1.2524 & 1.3501 & 3.0402 & 5.6844 & 0.0031 & 0.0042 & 0.0036 & 0.0081 & 0.0209 & 0.0234 & 0.0306 & 0.0734 \\
    & Gate &  11.9388 & 1.2863 &  5.4361 & 4.6863 & 0.0089 & 0.0043 &0.0066& 0.0064 & 0.0426  & 0.0607 & 0.0511& 0.0524 \\
    & Lego & 1.1857 & 1.5403 & 5.6895 & 5.1950 & 0.0024 & 0.0032 & 0.0073 & 0.0073 & 0.0243 & 0.0243 & 0.0546 & 0.0661 \\
    & Materials & 4.0039 & 1.5266 &9.3145& 4.1782 & 0.0043 & 0.0035 & 0.0151& 0.0088 & 0.0398 & 0.0177 &0.1096&0.0709\\
    & Snowtruck &  1.5929& 1.2136 &  8.6714 & 4.2423 & 0.0036 & 0.0032 &  0.0209 & 0.0058& 0.0267 & 0.0156 & 0.0659 & 0.0619\\
    & Statue &1.2508  & 1.2926  & 5.1592 &4.2392&0.0040 & 0.0032 & 0.0073 &0.0082 & 0.0184 & 0.0148 & 0.0539  &0.0654\\
    & Train &7.4066& 1.3244 & 10.9638 &4.4496& 0.0151 & 0.0030 & 0.0101 & 0.0069 & 0.0940 & 0.0139 & 0.0549 & 0.0588 \\
    \cline{2-14}
    & Mean & 3.7384 & 1.8775 & 6.8230 & 4.7167 & 0.0056 & 0.0043 & 0.0100 & 0.0072 & 0.0362 & 0.0263 & 0.0613 & 0.0630 \\
    \bottomrule
    \end{tabular}}
    \label{Table 7}
\end{table*}  

\begin{table*}[htbp]
    \caption{Quantitative rendering results of Mix-Step and MC-NeRF on all scenes. The comparison results show that our training sequence is effective, and simply stacking methods does not lead to better performance.}
    \centering
    \renewcommand\arraystretch{1.4}
        \resizebox{\textwidth}{!}{
    \begin{tabular}{c|c|cccc|cccc|cccc}
    \toprule
    & \multirow{2}{*}{Scene} & \multicolumn{4}{c|}{PSNR$\uparrow$} & \multicolumn{4}{c|}{SSIM$\uparrow$} & \multicolumn{4}{c}{LPIPS$\downarrow$}\\
    & & Ball & HalfBall & Room & Array & Ball & HalfBall & Room & Array & Ball & HalfBall & Room & Array \\
    \hline
    \multirow{9}{*}{Mix-Step} & Computer & 15.6289 & 14.1684 & 18.0732 & 22.0304 & 0.7937 & 0.8445 & 0.7438 & 0.9172 & 0.2376 & 0.2686 & 0.3228 & 0.0634 \\
    & Ficus & 17.4136 & 16.2707 & 21.0218 & 20.4416 & 0.9231 & 0.8959 & 0.9263 & 0.9479 & 0.1302 & 0.1752 & 0.1551 & 0.0941 \\
    & Gate & 19.0948 & 18.1299 & 23.5322 & 21.3187 & 0.9185 & 0.7781 & 0.9234 & 0.9358 & 0.1509 & 0.1874 & 0.1775 & 0.1045 \\
    & Lego & 20.7237 & 18.6574 & 21.0971 & 17.9275 & 0.8615 & 0.7870 & 0.7564 & 0.9099 & 0.1389 & 0.3503 & 0.2937 & 0.1658 \\
    & Materials & 18.6585 & 16.9294 & 24.4285 & 20.4265 & 0.9118 & 0.8764 & 0.9234 & 0.9531 & 0.1068 & 0.1386 & 0.0976 & 0.0753 \\
    & Snowtruck & 19.3878 & 20.0831 & 21.3667 & 16.7986 & 0.8891 & 0.9184 & 0.8259 & 0.9192 & 0.1417 & 0.0917 & 0.2615 & 0.1216 \\
    & Statue & 17.9503 & 19.1121 & 23.1452 & 17.5167 & 0.9231 & 0.8923 & 0.8876 & 0.9539 & 0.1161 & 0.1999 & 0.1767 & 0.0827 \\
    & Train & 18.6715 & 18.6559 & 18.8509 & 21.3719 & 0.8464 & 0.8810 & 0.9172 & 0.9175 & 0.1729 & 0.0966 & 0.2571 & 0.0875 \\
    \cline{2-14}
    & Mean & 18.4411 & 17.7508 & 21.4395 & 19.7290 & 0.8834 & 0.8592 & 0.8437 & 0.9318 & 0.1494 & 0.1885 & 0.2178 & 0.0994 \\
    \midrule
    \multirow{9}{*}{MC-NeRF} & Computer &  24.2311& 23.0962 & 23.6603 &24.6946 &  0.9217 &  0.8876 & 0.9008 & 0.9395 & 0.0794 & 0.0933 & 0.1004 & 0.0444 \\
    & Ficus &  24.9479 &  24.7662 &  26.5778 & 24.2929 & 0.9488 &  0.9070 &  0.9386 & 0.9504 & 0.0336 &  0.0519 &  0.0449 & 0.0387 \\
    & Gate & 27.9102  & 26.0876  & 29.3081 & 26.7734 & 0.9333  & 0.9131 & 0.9250 & 0.9356 & 0.0519  & 0.0525 & 0.0602 & 0.0439\\
    & Lego & 24.1226 & 23.4818 & 23.3858 & 24.3502 & 0.9227 & 0.8682 & 0.8837 & 0.9629 & 0.0620 & 0.0901 & 0.0874 & 0.0343 \\
    & Materials & 26.8742 & 26.6170 & 27.1323 & 25.0106 &  0.9484 & 0.9488 & 0.9529 & 0.9601 & 0.0321 & 0.0423 & 0.0294& 0.0397 \\
    & Snowtruck & 24.8742 & 24.0678 & 23.4415 & 23.3363 & 0.9211 & 0.8882 & 0.8937 & 0.9375 & 0.0433 & 0.0882 & 0.0874 & 0.0451 \\
    & Statue & 28.6882 & 29.2767 & 28.7953 & 24.6619 & 0.9731 & 0.9670 & 0.9612 & 0.9651 & 0.0303 & 0.0301 & 0.0398 & 0.0281\\
    & Train & 23.3830 & 23.5271 & 22.2576 & 23.4073 & 0.9311 & 0.8773 & 0.8589 & 0.9395 & 0.0454 & 0.0991 & 0.0835 & 0.0507\\
    \cline{2-14}
    & Mean & 25.6289 & 25.1151 & 25.5698 & 24.5659 & 0.9375 & 0.9072 & 0.9144 & 0.9488 & 0.0473 & 0.0684 & 0.0666 & 0.0406 \\
    \bottomrule
    \end{tabular}}
    \label{Table 8}
\end{table*}  

\begin{table*}[htbp]
    \caption{The Number of available images obtained by Different Methods. At the bottom of the table, $Total Number$ represents the total number of images in the current scene dataset, while the numbers at the top indicate the number of usable images after estimating camera poses using the respective methods. The table shows that in scenes with simple backgrounds or limited textures, SFM-based methods discard a large number of usable images, leading to reconstruction failures. In contrast, our method avoids this issue.}
    \centering
    \renewcommand\arraystretch{1.4}
        \resizebox{\textwidth}{!}{
    \begin{tabular}{c|cccc|cccc|cccc}
    \toprule
    \multirow{2}{*}{Scene} & \multicolumn{4}{c|}{COLMAP} & \multicolumn{4}{c|}{Meshroom} & \multicolumn{4}{c}{AprilTag}\\
    & Ball & HalfBall & Room & Array & Ball & HalfBall & Room & Array & Ball & HalfBall & Room & Array \\
    \hline
    Computer & 2 & 2 & 2 & 100 & 3 & 37 & 26 & 99 & 110 & 100 & 88 & 100 \\
    Ficus & 2 & 2 & 2 & 2 & 8 & 9 & 21 & 100 & 100 & 110 & 100 & 100 \\
    Gate & 16 & 31 & 31 & 98 & 41 & 67 & 34 & 100 & 110 & 100 & 88 & 100 \\
    Lego & 48 & 92 & 86 & 100 & 54 & 99 & 88 & 100 & 110 & 100 & 88 & 100 \\
    Materials & 38 & 2 & 71 & 5 & 36 & 0 & 36 & 0 & 110 & 100 & 88 & 100 \\
    Snowtruck & 33 & 96 & 28 & 100 & 63 & 97 & 84 & 100 & 110 & 100 & 88 & 100 \\
    Statue & 2 & 20 & 31 & 100 & 14 & 56 & 65 & 100 & 110 & 100 & 88 & 100 \\
    Train & 37 & 93 & 47 & 100 & 49 & 100 & 84 & 100 & 110 & 100 & 88 & 100 \\
    \cline{1-13}
    Total Number & 110 & 100 & 88 & 100 & 110 & 100 & 88 & 100 & 110 & 100 & 88 & 100 \\
    \bottomrule
    \end{tabular}}
    \label{Table 9}
\end{table*}

\begin{table*}[htbp]
    \caption{Comparison Rendering Results between Two SFM-based Composite Methods and MC-NeRF.}
    \centering
    \renewcommand\arraystretch{1.4}
        \resizebox{\textwidth}{!}{
    \begin{tabular}{c|c|cccc|cccc|cccc}
    \toprule
    & \multirow{2}{*}{Scene} & \multicolumn{4}{c|}{PSNR$\uparrow$} & \multicolumn{4}{c|}{SSIM$\uparrow$} & \multicolumn{4}{c}{LPIPS$\downarrow$}\\
    & & Ball & HalfBall & Room & Array & Ball & HalfBall & Room & Array & Ball & HalfBall & Room & Array \\
    \hline
    \multirow{9}{*}{COLMAP+NeRF} & Computer & - & - & - & 18.5081 & - & - & - & 0.8906 & - & - & - & 0.1677 \\
    & Ficus & - & - & - & - & - & - & - & - & - & - & - & - \\
    & Gate & - & 25.3121 & 21.1899 & 23.0474 & - & 0.9099 & 0.8889 & 0.9137 & - & 0.1206 & 0.2163 & 0.1195 \\
    & Lego & 20.8396 & 21.8068 & 22.2283 & 18.8225 & 0.9189 & 0.9074 & 0.8075 & 0.8749 & 0.1446 & 0.0494 & 0.1859 & 0.1936 \\
    & Materials & 22.5298 & - & 23.0754 & - & 0.8982 & - & 0.9577 & - & 0.1143 & - & 0.0957 & - \\
    & Snowtruck & 12.7945 & 26.4260 & - & 18.4196 & 0.7328 & 0.9155 & - & 0.8695 & 0.3671 & 0.0884 & - & 0.2072 \\
    & Statue & - & - & 22.1310 & 18.6054 & - & - & 0.9243 & 0.8903 & - & - & 0.1265 & 0.1205 \\
    & Train & 13.2663 & 27.0967 & 20.9158 & 18.4150 & 0.7214 & 0.9371 & 0.8816 & 0.8847 & 0.4036 & 0.0614 & 0.1205 & 0.1937 \\
    \cline{2-14}
    & Mean & 18.7213 & 25.1604 & 21.7981 & 19.4620 & 0.8499 & 0.9175 & 0.8920 & 0.8866 & 0.2087 & 0.0799 & 0.1490 & 0.1871 \\
    \midrule
    \multirow{9}{*}{Meshroom+NeRF} & Computer & - & - & - & 21.5518 & - & - & - & 0.9081 & - & - & - & 0.0978 \\
    & Ficus & - & - & - & 20.0217 & - & - & - & 0.9494 & - & - & - & 0.0990 \\
    & Gate & 17.8344 & 20.8989 & 22.219 & 21.3296 & 0.8807 & 0.8804 & 0.9110 & 0.9257 & 0.1852 & 0.2550 & 0.1642 & 0.2277 \\
    & Lego & 15.4833 & 18.3328 & 13.4810 & 22.2046 & 0.7851 & 0.7856 & 0.7227 & 0.9110 & 0.3341 & 0.2550 & 0.4191 & 0.1117 \\
    & Materials & 20.6267 & - & - & - & 0.9293 & - & - & - & 0.0998 & - & - & - \\
    & Snowtruck & 13.2374 & 16.9954 & 10.1618 & 19.4415 & 0.8998 & 0.8189 & 0.6761 & 0.9166 & 0.1758 & 0.2526 & 0.5043 & 0.1138 \\
    & Statue & - & 22.9819 & 14.1917 & - & - & 0.9684 & 0.8227 & - & - & 0.0547 & 0.2509 & - \\
    & Train & 21.1296 & 17.8849 & 11.8588 & 21.8684 & 0.8911 & 0.7979 & 0.6913 & 0.9237 & 0.1323 & 0.2526 & 0.4561 & 0.0997 \\
    \cline{2-14}
    & Mean & 16.7955 & 19.4188 & 14.3824 & 21.0696 & 0.8737 & 0.8502 & 0.7647 & 0.9224 & 0.1987 & 0.1920 & 0.3589 & 0.1249 \\
    \midrule
    \multirow{9}{*}{MC-NeRF} & Computer &  24.2311& 23.0962 & 23.6603 &24.6946 &  0.9217 &  0.8876 & 0.9008 & 0.9395 & 0.0794 & 0.0933 & 0.1004 & 0.0444 \\
    & Ficus &  24.9479 &  24.7662 &  26.5778 & 24.2929 & 0.9488 &  0.9070 &  0.9386 & 0.9504 & 0.0336 &  0.0519 &  0.0449 & 0.0387 \\
    & Gate & 27.9102  & 26.0876  & 29.3081 & 26.7734 & 0.9333  & 0.9131 & 0.9250 & 0.9356 & 0.0519  & 0.0525 & 0.0602 & 0.0439\\
    & Lego & 24.1226 & 23.4818 & 23.3858 & 24.3502 & 0.9227 & 0.8682 & 0.8837 & 0.9629 & 0.0620 & 0.0901 & 0.0874 & 0.0343 \\
    & Materials & 26.8742 & 26.6170 & 27.1323 & 25.0106 &  0.9484 & 0.9488 & 0.9529 & 0.9601 & 0.0321 & 0.0423 & 0.0294& 0.0397 \\
    & Snowtruck & 24.8742 & 24.0678 & 23.4415 & 23.3363 & 0.9211 & 0.8882 & 0.8937 & 0.9375 & 0.0433 & 0.0882 & 0.0874 & 0.0451 \\
    & Statue & 28.6882 & 29.2767 & 28.7953 & 24.6619 & 0.9731 & 0.9670 & 0.9612 & 0.9651 & 0.0303 & 0.0301 & 0.0398 & 0.0281\\
    & Train & 23.3830 & 23.5271 & 22.2576 & 23.4073 & 0.9311 & 0.8773 & 0.8589 & 0.9395 & 0.0454 & 0.0991 & 0.0835 & 0.0507\\
    \cline{2-14}
    & Mean & 25.6289 & 25.1151 & 25.5698 & 24.5659 & 0.9375 & 0.9072 & 0.9144 & 0.9488 & 0.0473 & 0.0684 & 0.0666 & 0.0406 \\
    \bottomrule
    \end{tabular}}
    \label{Table 10}
\end{table*}  

\begin{table*}[htbp]
\caption{Quantitative rendering results of SUPPLEMENTARY EXPERIMENT E.}
    \centering
    \renewcommand\arraystretch{1.3}
    \resizebox{0.48\textwidth}{!}{
    \begin{tabular}{ccccccc}
    \toprule
      &  & $40^{\circ}$ & $50^{\circ}$ & $60^{\circ}$ & $70^{\circ}$ & $80^{\circ}$
    \\\hline
    \multirow{3}{*}{PSNR$\uparrow$} & Fixed FOV & 25.440 & 27.872 & 29.896 & 32.935 & 34.783 \\
     &Mixed FOV&24.169&27.333&29.949&32.138&33.786 \\
     &Distance&1.271&0.539&0.053&0.797&0.997 \\
    \hline
    \multirow{3}{*}{SSIM$\uparrow$} & Fixed FOV & 0.858 & 0.913 & 0.942 & 0.969 & 0.980 \\
     &Mixed FOV&0.819&0.899&0.939&0.963&0.975 \\
     &Distance&0.039&0.014&0.003&0.006&0.005 \\
     \hline
    \multirow{3}{*}{LPIPS$\downarrow$} & Fixed FOV & 0.108 & 0.056 & 0.040 & 0.017 & 0.014 \\
     &Mixed FOV&0.147&0.069&0.037&0.026&0.016 \\
     &Distance&0.039&0.013&0.003&0.009&0.002 \\
    \bottomrule
    \end{tabular}}
    \label{Table 11}
\end{table*}

\end{document}